\documentclass[11pt]{article}
\usepackage{makecell}
\usepackage[preprint]{acl}
\usepackage{booktabs}
\usepackage{amsmath}
\usepackage{mathtools}
\usepackage{multirow}
\usepackage{caption} 
\usepackage{times}
\usepackage{latexsym}
\usepackage[table]{xcolor}
\usepackage[breakable,skins]{tcolorbox}
\newcommand*\justify{%
  \fontdimen2\font=0.4em% interword space
  \fontdimen3\font=0.2em% interword stretch
  \fontdimen4\font=0.1em% interword shrink
  \fontdimen7\font=0.1em% extra space
  \hyphenchar\font=`\-% allowing hyphenation
}

\renewcommand{\texttt}[1]{%
\begingroup
\ttfamily
\begingroup\lccode`~=`/\lowercase{\endgroup\def~}{/\discretionary{}{}{}}%
\begingroup\lccode`~=`[\lowercase{\endgroup\def~}{[\discretionary{}{}{}}%
\begingroup\lccode`~=`.\lowercase{\endgroup\def~}{.\discretionary{}{}{}}%
\catcode`/=\active\catcode`[=\active\catcode`.=\active
\justify\scantokens{#1\noexpand}%
\endgroup
}

\PassOptionsToPackage{numbers, compress}{natbib}

\usepackage[utf8]{inputenc}         %
\usepackage[T1]{fontenc}            %
\usepackage{graphicx}
\usepackage{todonotes}
\usepackage{multirow}
\usepackage{hyperref}
\usepackage[capitalize,noabbrev,nameinlink,sort]{cleveref}
\usepackage{subcaption}
\usepackage{multicol}
\usepackage{array}
\usepackage{enumitem}
\usepackage{algorithm}
\usepackage{algpseudocode}
\usepackage{tabularx}
\usepackage{listings}
\usepackage{makecell}
\usepackage{xspace}
\usepackage{colortbl}
\usepackage{fontawesome5}
\usepackage{pifont}
\usepackage[normalem]{ulem}
\useunder{\uline}{\ul}{}
\usepackage{tablefootnote}
\usepackage{tcolorbox}
\usepackage{arydshln}
\usepackage[toc, page, header]{appendix}
\usepackage{tabularx}
\usepackage{listings}

\usepackage{amsmath}
\usepackage{amssymb}
\usepackage{amsfonts}                               % blackboard math symbols
\usepackage{amsthm}
\usepackage{mathrsfs}
\usepackage{bm}                                     % bm command
\usepackage{blkarray}                               % to support matrix
\usepackage{nicefrac}                               % compact symbols for 1/2, etc.
\usepackage{bbm}

\usepackage{wrapfig}
\usepackage{graphicx}                               % include pdf figures
\usepackage{caption}
\usepackage{tikz}
\usepackage{circuitikz}
\usetikzlibrary{patterns,snakes}
\usetikzlibrary{positioning,calc,fit,decorations.pathmorphing,shapes.geometric, shapes.gates.logic.US, calc}
\usetikzlibrary{arrows,arrows.meta,decorations.markings,shapes,shapes.arrows}
\usetikzlibrary{decorations,decorations.pathreplacing}
\usetikzlibrary{backgrounds}
\usepackage{filecontents}                           % support to pgfplots
\usepackage{pgfplots}
\usepackage{pgfplotstable}
\usepgfplotslibrary{groupplots}
\usepackage{scalefnt}
\pgfplotsset{compat=newest}
\usepackage{xcolor}

\definecolor{firstcolor}{HTML}{C3423F}
\definecolor{secondcolor}{HTML}{2A4B8C}
\definecolor{aworld_blue}{HTML}{4e81ff}
\definecolor{aworld_cyan}{HTML}{41d7fa}
\definecolor{aworld_teal}{HTML}{5fede4}
\definecolor{coral}{RGB}{255,127,80}
\definecolor{darkgreen}{RGB}{0,100,0}
\definecolor{darkyellow}{RGB}{204,153,0}
\definecolor{salmon}{RGB}{250,128,114}
\definecolor{darkred}{RGB}{150,0,0}

\hypersetup{
  colorlinks=true,
  linkbordercolor=aworld_blue,
  citebordercolor=aworld_cyan,
  urlbordercolor=aworld_teal
}

\definecolor{improvementblue}{RGB}{55,126,184}    % Blue (#377eb8)
\definecolor{degradationorange}{RGB}{230,85,13}   % Orange (#e6550d)

\def\eqref#1{equation~\ref{#1}}
\def\1{\bm{1}}

\DeclareMathAlphabet{\mathsfit}{\encodingdefault}{\sfdefault}{m}{sl}
\SetMathAlphabet{\mathsfit}{bold}{\encodingdefault}{\sfdefault}{bx}{n}

\usepackage[T1]{fontenc}
\DeclareTextSymbolDefault{\textquotedbl}{T1}
\usepackage{pifont}
\usepackage{bbm}

\usepackage[utf8]{inputenc}
\usepackage{subcaption}
\usepackage{upquote}  % straight quotes/backticks in verbatim and listings
\usepackage{microtype}
\usepackage{bbding}
\usepackage{inconsolata}
\usepackage{graphicx}

\title{}
\author{}

\begin{document}

% Full-width title block before the two-column body.
\makeatletter
\twocolumn[{%
\begin{@twocolumnfalse}
% Top logo: one fixed-ratio lockup matching the original logo/pdf geometry.
\noindent\includegraphics[height=1.0cm]{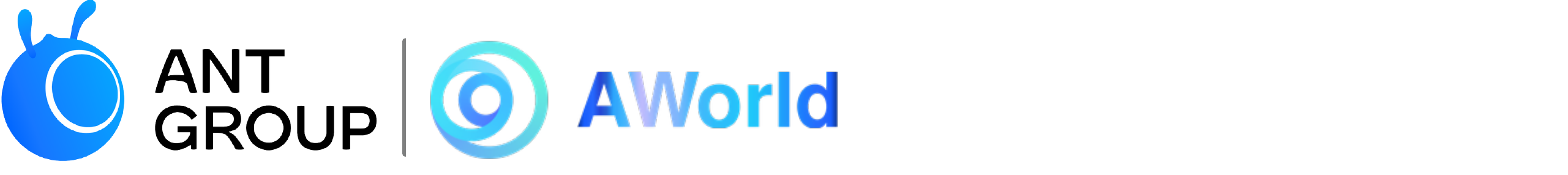}\\[0.2em]
\noindent\rule{\textwidth}{0.4pt}

% Manually typeset title.
\vspace{0.25em}
\begin{center}
{\Large\textbf{Design the Feedback, Not Just the Reward:}}\\[0.25em]
{\Large\textbf{Environments for Open-Ended Reinforcement Learning}}
\end{center}
\vspace{0.05em}

% Rule below the title.
\noindent\rule{\textwidth}{0.4pt}

% Manually typeset author block.
\vspace{0.3em}
\begin{center}
\textbf{Rui Jin\textsuperscript{1,2*}}, \textbf{Jikai Chen\textsuperscript{1*}},
\\[0.2em]
\textbf{Chengyue Yu\textsuperscript{1}},
\textbf{Yihan Chen\textsuperscript{1,3}},
\textbf{Hao Zhou\textsuperscript{1,3}},
\textbf{Demin Zhu\textsuperscript{1,4}},
\\[0.2em]
\textbf{Kaichen Yang\textsuperscript{1,5}},
\textbf{Dong Wang\textsuperscript{1}},
\textbf{Linjian Mo\textsuperscript{1}},
\\[0.2em]
\textbf{Chenyi Zhuang\textsuperscript{1$\ddag$}}
\\[0.35em]
\small
\textsuperscript{1}Inclusion AI, Ant Group \quad
\textsuperscript{2}Zhongnan University \quad
\\[0.1em]
\textsuperscript{3}Zhejiang University \quad
\textsuperscript{4}Central University of Finance and Economics
\\[0.1em]
\textsuperscript{5}Dalian University of Technology
\end{center}

\vspace{0.25em}

% Disabled: first-page hero figure from the previous version (kept for reference).
% % Main result figure.
% \noindent
% \begin{minipage}{\textwidth}
%   \centering
%   \includegraphics[width=\textwidth]{img/pass_rate_chart.png}
%   \vspace{-1.0em}
%   \captionof{figure}{\textbf{Overall model pass rate on MiniAppBench.}
%   Average pass rate comparison across representative open-source and closed-source baselines. Ling-RCCA-Flash is highlighted as our RCCA-trained model.}
%   \label{fig:main_results_charts}
% \end{minipage}
%
% \vspace{0.1em}
\end{@twocolumnfalse}
}]
\makeatother

% Footnote for equal contributions and corresponding authors
{\renewcommand\thefootnote{}
\footnotetext{*Equal contributions. $\ddag$Corresponding Authors.}}

% !TeX root = ../main.tex

\begin{abstract}
\setlength{\parskip}{0.25em}
Reinforcement learning has advanced large language models on tasks with reliable verifiers, where checking an outcome provides a clear basis for reward. Extending this approach to open-ended tasks requires environments that can both evaluate complex, partially successful outputs and turn these judgments into effective learning guidance. We propose a reinforcement learning framework built around a training environment that integrates human domain knowledge with model reasoning. Human domain knowledge, encoded in rubrics, anchors evaluation to task intent, while model reasoning combined with hierarchical rewards distinguishes complex open-ended outputs and links diagnosed failures to likely responsible output regions. Furthermore, these diagnostic attributions guide token-level weights in the GRPO objective, thereby translating the ambiguous notion of success in open-ended tasks into concrete guidance for policy optimization. We instantiate this framework in interactive web application generation, a representative open-ended generation task. The resulting model achieves an average pass rate of 41.25\% on MiniAppBench and 76.19\% on ArtifactsBench. Ablation studies show the contributions of individual components, while additional out-of-distribution evaluations support transferable gains without degrading performance on other evaluated tasks.
\end{abstract}
% !TeX root = ../main.tex

\section{Introduction}

\begin{figure*}[t]
    \centering
    \includegraphics[width=\textwidth]{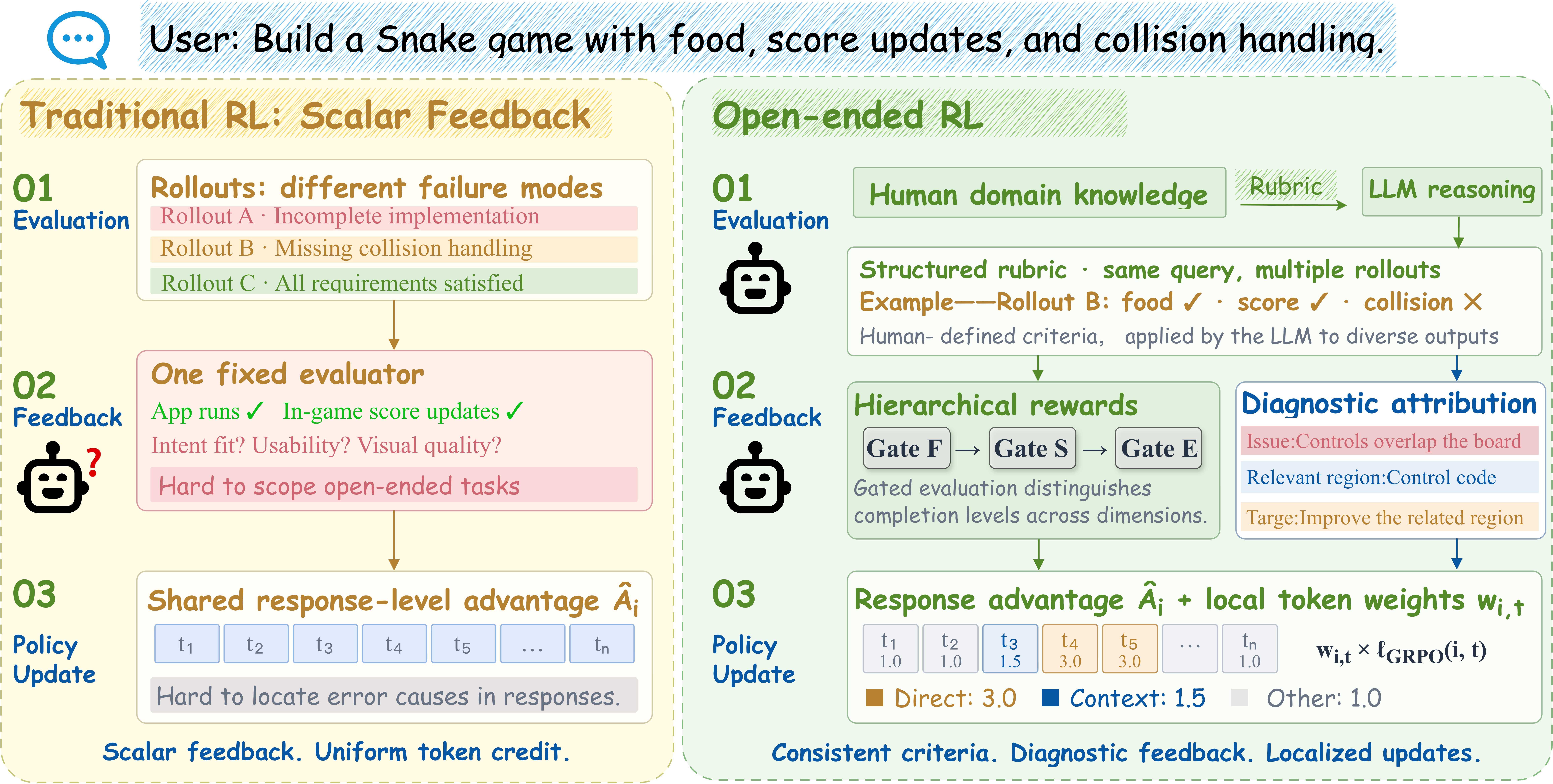}
    \caption{\textbf{Connecting environment judgments to policy updates.}
    A scalar-only training interface aggregates evaluation outcomes into a reward and yields a shared response-level advantage. Our environment also preserves diagnostic evidence, links observed failures to relevant output regions, and uses these attributions to guide token-level optimization weights.}
    \label{fig:intro_motivation}
\end{figure*}

Reinforcement learning has become an important mechanism for improving large language models (LLMs), with substantial progress on mathematical reasoning and competitive programming~\cite{deepseekmath,deepseekr1}. Reliable verifiers provide a foundation for this progress: reference answers or executable tests allow task outcomes to be checked and converted into rewards. As LLM applications expand to open-ended tasks, including application generation, complex workflows, and multi-step agentic problems~\cite{aworld,webcoderbench}, training environments must support more varied forms of evaluation.

These properties complicate both evaluation and learning. Assessing open-ended responses involves judgments about quality whose underlying criteria are often ambiguous and left implicit in the user query. Evaluation also spans multiple dimensions, making it difficult to translate these judgments into rewards that clearly distinguish different degrees of task fulfillment. Furthermore, even when the environment identifies a specific failure, an overall reward does not explicitly indicate which parts of the generated output contributed to it. These challenges motivate a central question for open-ended reinforcement learning: how can a training environment establish clear evaluation criteria, differentiate complex task outcomes, and turn diagnostic feedback into effective guidance for policy updates?

We propose a reinforcement learning framework built around a training environment that integrates human domain knowledge with model reasoning. Because an open-ended task can be fulfilled in different ways, reliable evaluation must maintain consistent success criteria while adapting its judgments to each output. We therefore encode human domain knowledge in rubrics to establish task requirements and priorities, while the environment uses LLM reasoning to assess how different outputs satisfy these requirements. To distinguish sampled outputs across multiple evaluation dimensions, we use hierarchical rewards to organize the resulting judgments into differentiated signals for group-relative learning. The environment also produces requirement-level diagnoses that link observed failures to likely responsible output regions. Together, these mechanisms keep evaluation anchored to task intent while accommodating diverse realizations, providing both meaningful quality comparisons and local evidence to guide policy optimization.

Furthermore, these diagnostic attributions guide how the policy learns from its outputs. Standard GRPO computes a response-level advantage from scalar rewards and shares that advantage across generated tokens~\cite{deepseekr1}. Our framework uses the environment's diagnostic evidence to determine token-level weights in the GRPO objective, modulating the response-level learning signal according to the attributed responsibility of different output regions. Overall rewards continue to support comparisons among responses, while local weights make the environment's explanations relevant to updates within each response. By grounding evaluation in human intent and using diagnostic attributions to weight token-level terms in the GRPO objective, we establish concrete optimization targets for open-ended tasks whose success criteria are ambiguous and whose failures are difficult to localize. Figure~\ref{fig:intro_motivation} illustrates the complete training environment.

We instantiate the framework in interactive web application generation, a representative open-ended generation task in which an LLM produces complete HTML, CSS, and JavaScript from a natural-language request~\cite{design2code,miniapp,artifactsbench}. This setting embodies the challenges discussed above. The same request admits diverse implementations, and judging their quality requires interpreting user intent across functionality, interaction, and presentation. An application may appear correct while failing important behavioral requirements, making overall quality difficult to capture with a single undifferentiated assessment. Furthermore, a user-visible failure does not directly reveal which implementation choices caused it. Source code and observable browser behavior provide concrete evidence for investigating these connections, making this setting suitable for studying how an environment can turn complex task judgments into localized learning guidance.

In this setting, we define functional rubrics that specify the behaviors an application should implement and the criteria for judging their correctness. These rubrics guide both training-task synthesis and application evaluation. For each generated application, the environment checks output format, source-code validity, and runtime validity before assessing rubric-level functionality, producing differentiated reward signals for group-relative learning. The functional assessment also provides the basis for local credit assignment: using interaction results and source code, the environment diagnoses unmet requirements and identifies likely responsible code regions, such as event handlers, state updates, DOM fragments, and CSS selectors. These attributed spans are then aligned with generated tokens to determine weights in the GRPO objective. In this way, the same evaluation process supplies both application-level rewards and requirement-level diagnoses that guide updates to relevant implementation regions.

The resulting model achieves an average pass rate of \textbf{41.25\%} on MiniAppBench and a state-of-the-art score of \textbf{76.19} on ArtifactsBench, two benchmarks for evaluating interactive web application generation. These results improve upon its supervised initialization by \textbf{14.40} percentage points and \textbf{4.48} points, respectively. Ablation studies show the contributions of individual components. Further evaluations on out-of-distribution tasks suggest that the proposed training paradigm helps the model acquire generalizable knowledge and skills, extending the benefits of training beyond the target benchmarks.

Our contributions are summarized as follows:
\begin{itemize}[leftmargin=*,topsep=2pt,itemsep=1pt,parsep=0pt,partopsep=0pt]
    \item We propose a reinforcement learning framework for open-ended tasks built around a training environment that integrates human domain knowledge with model reasoning, connecting task specification, evidence-grounded evaluation, and policy updates.
    \item We develop hierarchical rewards and diagnostic attribution within this environment to distinguish multidimensional task outcomes and translate requirement-level diagnoses into token-level optimization weights for GRPO.
    \item We instantiate the framework in interactive web application generation, achieving gains on MiniAppBench and ArtifactsBench, with ablation studies demonstrating the contributions of individual components and out-of-distribution evaluations suggesting the acquisition of generalizable knowledge and skills.
\end{itemize}

\section{Related Work}

\paragraph{Reinforcement learning for open-ended tasks.}
Reinforcement learning has achieved substantial progress on tasks with reliable verifiers, such as mathematical reasoning and code generation, where reference answers or executable tests can directly determine correctness \cite{deepseekmath,deepseekr1}. Extending RL to open-ended tasks is more challenging because success is often ambiguous, multidimensional, and cannot be captured by a single deterministic criterion. Unlike verifier-based settings, open-ended outputs may admit multiple valid realizations while only partially satisfying different aspects of a request. Evaluation must therefore preserve distinctions among validity, task completion, and presentation quality rather than collapse heterogeneous failures into a common outcome. This shifts the design question from whether a final response is globally acceptable to what evidence an environment should retain so that optimization can compare diverse responses while remaining aligned with human intent. Recent work has therefore explored LLM-based evaluation and rubric-guided judging for complex and open-ended generation \cite{liu2023geval,kim2024prometheus}. These approaches aim to make implicit task requirements explicit and distinguish different degrees of task fulfillment, providing richer supervision than a single holistic score.

\paragraph{Fine-grained feedback and credit assignment.}
Most language-model RL methods, including PPO-style RLHF and GRPO, rely primarily on sequence- or trajectory-level rewards \cite{schulman2017ppo,ouyang2022training,deepseekmath}. To obtain more informative learning signals, prior work has explored process supervision, step-level rewards, segment-level feedback, and finer-grained credit assignment \cite{lightman2024verify,wang2024mathshepherd,guo2025spo}. Natural-language feedback has also been used for critique, revision, and iterative policy improvement \cite{akyurek2023rl4f,shinn2023reflexion}. These directions expose a persistent mismatch between feedback resolution and optimization resolution: a response-level score can compare complete outputs but assigns the same learning signal to every token, whereas a local critique identifies a defect without specifying how it should interact with the response's relative quality. Effective credit assignment for long, structured outputs therefore requires both levels. The global signal should preserve policy-level comparisons, while localized evidence should determine where that signal is emphasized within each response. Interactive web application generation provides a representative open-ended setting for studying these problems, as evaluation must jointly consider functionality, visual presentation, and interaction behavior. Benchmarks such as WebDev Arena, ArtifactsBench, and MiniAppBench increasingly incorporate human preference, structured requirements, and dynamic interaction into application-level evaluation \cite{vichare2025webdev,artifactsbench,miniapp}.

% !TeX root = ../main.tex
\section{Method}

\subsection{Training Environment and Problem Formulation}

We formulate open-ended reinforcement learning around a training environment that connects task specification, evaluation, and policy optimization. Let $x$ denote the expanded task specification derived from an original user query. Given $x$, a policy $\pi_\theta$ generates an output $y$. Each task is associated with a rubric set
\[
\mathcal{R}=\{r_1,\ldots,r_M\},
\]
which specifies the requirements against which the output is assessed. Human domain knowledge, encoded in rubrics developed with LLM assistance, establishes consistent evaluation criteria, while LLM reasoning within the environment assesses how individual outputs satisfy them. This separates the criteria for success from the particular form of a valid solution.

\begin{figure*}[t]
	\centering
	\includegraphics[width=\textwidth]{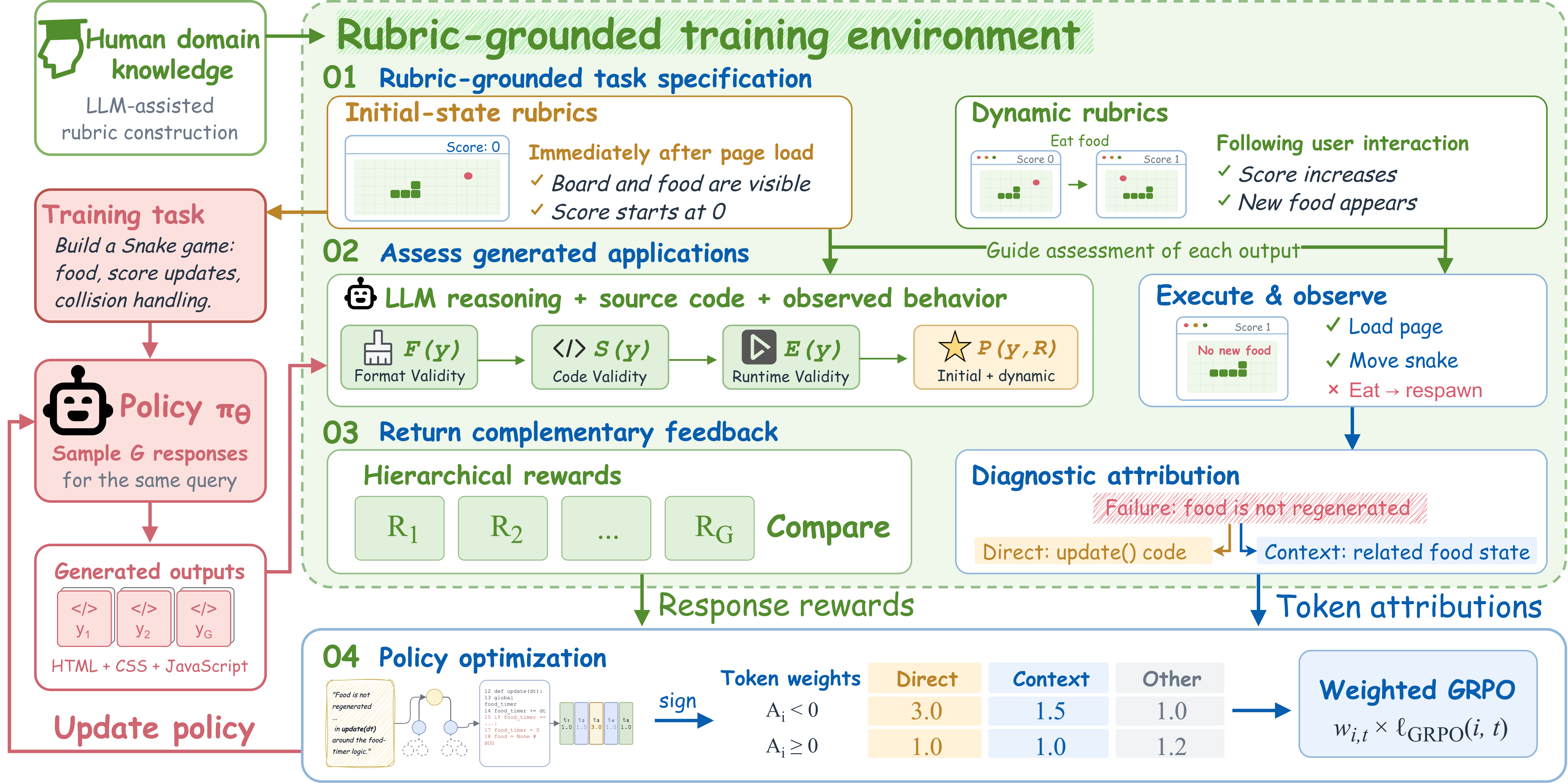}
	\caption{\textbf{Overview of the proposed training environment.}
		Human domain knowledge, expressed through rubrics, guides the environment's assessment of generated outputs. Hierarchical rewards support comparisons among responses, while diagnostic attributions connect failures to relevant output regions and determine token-level weights for policy optimization.}
	\label{fig:method_overview}
\end{figure*}

The environment produces two complementary forms of feedback:
\[
\mathcal{E}(x,y,\mathcal{R})
=\bigl(R(y,\mathcal{R}),\mathcal{D}(y)\bigr),
\]
where $R(y,\mathcal{R})$ is a scalar reward derived from hierarchical evaluation gates that distinguish responses by their validity and degree of task fulfillment, providing differentiated signals for group-relative learning. Diagnostic attribution $\mathcal{D}(y)$ further links detected failures to relevant output regions, allowing targeted updates within complex open-ended outputs. Figure~\ref{fig:method_overview} summarizes this training process.

We instantiate this environment in interactive web application generation, where $y$ is a complete HTML, CSS, and JavaScript application. Rubrics specify user-facing requirements, and the environment assesses generated applications using source code and code-inferred execution behavior. The following sections describe how this instantiation defines tasks, constructs rewards and diagnoses, and converts diagnostic evidence into optimization weights.

\subsection{Rubric-Grounded Task Specification}
\label{sec:rubrics}

To make open-ended application quality assessable without prescribing a single implementation, we express task intent as functional rubrics. Human domain knowledge, supported by LLM-assisted rubric construction, defines the expected behaviors and evaluation priorities. These rubrics guide training-task synthesis, yielding examples of the form $(x,\mathcal{R})$. Each rubric describes an independently checkable user-facing requirement, providing a concrete evaluation unit while allowing different implementations to satisfy it.

We distinguish two types of requirements. \textit{Initial-state rubrics} specify properties that should hold immediately after page load, such as required interface elements, default content, or initial values. \textit{Dynamic rubrics} specify behavior following user interaction, such as opening a panel after a click, updating displayed content from an input field, or performing an expected state transition. The environment assesses initial-state requirements from the page-load structure implied by the generated source code and assesses dynamic requirements from the implementation of the corresponding interaction logic.

For each task, the rubric serves as the shared basis for task construction, reward computation, and requirement-level diagnosis. Thus, the requirements used to define a task also determine how its outputs are assessed and how functional failures are traced to the implementation.

\subsection{Hierarchical Evaluation and Rewards}
\label{sec:rewards}

Open-ended outputs differ along multiple dimensions, so directly aggregating requirement-level outcomes can obscure distinctions that matter for group-relative learning. In web application generation, a response may have an invalid format, contain source-code errors, fail during execution, or execute successfully while violating functional requirements. The environment therefore organizes assessment into four stages: \emph{(1) output-format validity}, \emph{(2) source-code validity}, \emph{(3) runtime validity}, and \emph{(4) rubric-level requirement satisfaction}.

The environment first extracts the final HTML application and checks its output format. It then inspects the HTML, CSS, and JavaScript for source-level errors and checks whether the implementation is executable according to source-code diagnostics. Applications that pass these validity checks proceed to rubric-level assessment of their initial-state and interaction requirements. This hierarchy distinguishes validity and execution failures from different degrees of requirement satisfaction, providing differentiated rewards for group-relative learning.

Let $F(y)$, $S(y)$, and $E(y)$ indicate whether $y$ passes format, source-code, and runtime validation, respectively. The gated reward is
\[
R(y,\mathcal{R})=
\begin{cases}
	0, & F(y)=0,\\
	0, & F(y)=1,\ S(y)=0,\\
	0.1, & F(y)=S(y)=1,\\
	& \text{and } E(y)=0,\\
	R_{\mathrm{rubric}}, & F(y)=S(y)=1,\\
	& \text{and } E(y)=1.
\end{cases}
\]
Here, $R_{\mathrm{rubric}}\in(0.2,1.0]$. Format and source-code failures share the lowest reward, runtime failures receive an intermediate reward, and applications that reach rubric-level assessment receive a higher reward determined by requirement satisfaction.

For applications that pass the validity gates, the environment further evaluates the generated application against the rubric set $\mathcal{R}$. Let $\mathcal{V}(y,\mathcal{R})$ denote the set of violated rubrics after retaining at most one highest-severity finding for each rubric item. The rubric-level penalty is computed by aggregating the per-item penalties as
\[
P(y,\mathcal{R})
=
\min\left\{
\sum_{r_j\in\mathcal{V}(y,\mathcal{R})}
p_j,\,
1-\epsilon_R
\right\},
\]
where $p_j \geq 0$ denotes the penalty associated with the violation of rubric $r_j$, determined by the item's functional level and violation severity, and $\epsilon_R>0$ is a small constant used to keep the rubric reward strictly above the runtime-failure reward. The detailed penalty assignment is provided in Appendix~\ref{app:reward-details}. The clipping operation ensures that
$
1-P(y,\mathcal{R})>0.
$
The rubric-level reward is then computed as
\[
R_{\mathrm{rubric}}
=
0.2+0.8\left(1-P(y,\mathcal{R})\right).
\]

\subsection{Diagnostic Attribution to Generated Outputs}

Alongside the scalar reward, the environment also provides extra diagnostic feedback to support fine-grained credit assignment within each response. Guided by the task rubrics defined in Section~\ref{sec:rubrics}, it identifies unmet functional requirements and highlights potentially responsible code fragments. We denote the collected diagnostic findings by
\[
\mathcal{D}(y)=\{d_1,\ldots,d_K\},
\]
where each $d_k$ describes a detected failure and records supporting implementation evidence, such as relevant code excerpts, event handlers, or state variables.

The environment uses this evidence to identify code regions likely responsible for each issue. Because an interactive behavior can depend on several related implementation regions, a directly identified fragment may not cover all relevant code. We therefore expand the initial locations through source-level relations, including enclosing functions, event bindings, state reads and writes, and function calls.

For issue $d_k$, let $\mathcal{S}_k^{\mathrm{dir}}$ denote source spans directly implicated by the diagnosis and $\mathcal{S}_k^{\mathrm{ctx}}$ denote additional spans related through this expansion. For each response $y_i$, we collect the directly implicated and contextually related code spans across all diagnostic findings and map them to the corresponding generated tokens. This yields token sets $\mathcal{T}_i^{\mathrm{dir}}$ and $\mathcal{T}_i^{\mathrm{ctx}}$, respectively, which are used to construct the token-level optimization weights described next.

\subsection{Policy Optimization with Diagnostic Weights}

For each query $x$, we sample a group of $G$ responses and compute the standard GRPO advantage $A_i$ from their hierarchical rewards described in Section~\ref{sec:rewards}. Standard GRPO applies this response-level advantage across all tokens in $y_i$. Our framework retains the reward and advantage computation while using diagnostic attribution to determine the relative emphasis of token-level terms.

We assign a diagnostic weight $\bar{w}_{i,t}$ of $3.0$ to tokens directly implicated in an issue, $1.5$ to tokens associated only with contextual spans, and $1.0$ to all remaining tokens. Because GRPO advantages reflect relative quality within a group, a response with diagnosed errors can still receive a positive advantage. Increasing the weights of these error regions would then amplify their positive updates. We therefore apply diagnostic weights according to the sign of the response-level advantage. For a negative-advantage response, diagnosed regions receive stronger negative updates. A positive-advantage response can still contain errors, so its diagnosed regions retain the baseline weight, while regions outside attributed spans receive a modestly larger positive weight.

Accordingly, the effective token weight is
\[
w_{i,t}
=
\begin{cases}
	\bar{w}_{i,t}, & A_i<0,\\
	1, & A_i\geq0\ \text{and}\ 
	t\in\mathcal{T}_i^{\mathrm{dir}}\cup\mathcal{T}_i^{\mathrm{ctx}},\\
	c, & A_i\geq0\ \text{and}\ 
	t\notin\mathcal{T}_i^{\mathrm{dir}}\cup\mathcal{T}_i^{\mathrm{ctx}},
\end{cases}
\]
where $c=1.2$. We incorporate these weights into the clipped GRPO objective:
\[
\begin{aligned}
&\mathcal{L}_{\mathrm{policy}}
= -\frac{1}{G}\sum_{i=1}^{G}\frac{1}{T_i}\sum_{t=1}^{T_i}\\
&\min\!\bigl(\rho_{i,t}w_{i,t}A_i,\\
&\operatorname{clip}(\rho_{i,t},1-\epsilon_{\mathrm{low}},\\
&\qquad\; 1+\epsilon_{\mathrm{high}})\,w_{i,t}A_i
\bigr).
\end{aligned}
\]
\[
\rho_{i,t}
=
\frac{
	\pi_\theta(y_{i,t}\mid x,y_{i,<t})
}{
	\pi_{\theta_{\mathrm{old}}}(y_{i,t}\mid x,y_{i,<t})
},
\]
where $T_i$ is the length of response $y_i$, $\rho_{i,t}$ is the probability ratio between the current policy and the policy used to sample the responses, and we use $\epsilon_{\mathrm{low}}=0.2$ and $\epsilon_{\mathrm{high}}=0.28$. The diagnostic weights adjust the relative strength of token-level updates while retaining the standard GRPO advantage computation and clipping mechanism.

Through this objective, the environment's judgments serve two connected purposes: hierarchical rewards determine response-level preferences, and diagnostic attributions determine where those preferences receive greater emphasis within each response. Human-grounded task criteria, model-based assessment, and token-level optimization thus form a single training process for learning from open-ended outputs.
% !TeX root = ../main.tex

\section{Experiments}
\label{sec:experiments}
\subsection{Experimental Setup}
\label{sec:experimental-setup}

We instantiate the framework on Ling-3.0-Flash, a 124B-parameter hybrid-linear mixture-of-experts model with 5.1B activated parameters per token~\cite{ling3flash}. Starting from the base model, we first perform supervised fine-tuning (SFT) for interactive web application generation. We then apply reinforcement learning using the proposed environment, which evaluates generated applications against task rubrics, constructs hierarchical rewards, and converts diagnostic attributions into token-level optimization weights. We denote the resulting model as \textbf{Ling-3.0-Flash + RL (Ours)}.

We evaluate the base model, its SFT initialization, and the final RL model. MiniAppBench~\cite{miniapp} serves as the primary benchmark, measuring interactive HTML application generation across three difficulty levels and six domains. MiniAppBench results are averaged over eight evaluation runs. We additionally evaluate on ArtifactsBench~\cite{artifactsbench}, which uses a different task distribution and evaluation protocol for visual-interactive artifacts. We further conduct ablation studies to examine the contributions of hierarchical rewards and diagnostic token weighting, and analyze how local error attribution interacts with group-relative learning. All RL variants start from the same SFT initialization. Additional benchmark evaluations assess whether the resulting gains transfer to code editing, instruction following, knowledge reliability, and reasoning.

\subsection{Main Results}
\label{sec:main-results}

\begin{table*}[t]
\centering
\caption{\textbf{Main results on MiniAppBench.}
Pass rates are reported by difficulty and domain, together with the overall average. Token consumption and inference time are included when available.}
\label{tab:miniapp-main}
\footnotesize
\setlength{\tabcolsep}{3.5pt}
\setlength{\abovecaptionskip}{3pt}
\setlength{\belowcaptionskip}{2pt}
\renewcommand{\arraystretch}{1.04}
\providecommand{\modelicon}[1]{\raisebox{-0.24em}{\includegraphics[height=1.45em]{img/logo_official_png/#1.png}}}
\providecommand{\anticon}{\raisebox{-0.24em}{\includegraphics[height=1.45em]{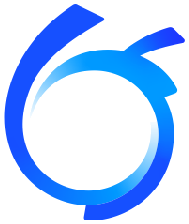}}}
\resizebox{\textwidth}{!}{%
\begin{tabular}{clcccccccccccc}
\toprule
\multicolumn{2}{c}{\multirow{3}{*}{Model}} & \multicolumn{9}{c}{Pass Rate (\%)} & \multirow{3}{*}{Avg. (\%)} & \multirow{3}{*}{Tokens} & \multirow{3}{*}{Time (s)} \\
\cmidrule(lr){3-11}
\multicolumn{2}{c}{} & \multicolumn{3}{c}{Difficulty} & \multicolumn{6}{c}{Domain} & & & \\
\cmidrule(lr){3-5}\cmidrule(lr){6-11}
\multicolumn{2}{c}{} & Easy & Mid & Hard & Games & Science & Tools & Humanities & Viz. & Lifestyle & & & \\
\midrule
\rowcolor{blue!10}
\multicolumn{14}{c}{\textbf{Reference Models}} \\
\modelicon{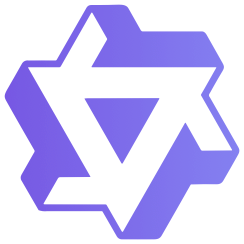} & Qwen3-32B & 1.59 & 0.55 & 0.00 & 0.00 & 0.57 & 0.00 & 0.00 & 2.04 & 3.70 & 0.66 & 3,470.68 & 22.16 \\
\modelicon{qwen} & Qwen3-235B-A22B & 6.43 & 2.35 & 0.00 & 0.93 & 0.60 & 4.00 & 4.88 & 7.27 & 10.34 & 2.88 & 4,068.27 & 49.55 \\
\modelicon{qwen} & Qwen3-Coder-480B-A35B-Instruct & 6.06 & 0.00 & 0.00 & 0.00 & 0.00 & 0.00 & 0.00 & 9.43 & 11.11 & 1.83 & 2,324.83 & 25.04 \\
\modelicon{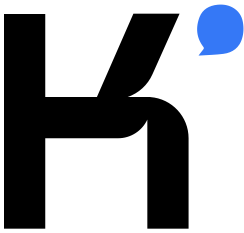} & Kimi-K2-Instruct & 14.17 & 5.03 & 0.00 & 3.77 & 3.11 & 4.08 & 4.88 & 17.65 & 18.52 & 6.19 & 3,435.97 & 46.76 \\
\modelicon{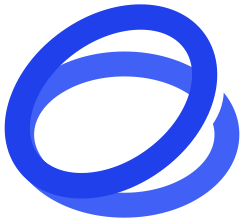} & GLM-5.1 & 41.50 & 31.40 & 28.50 & 23.70 & 27.70 & 32.10 & 48.60 & 59.60 & 33.30 & 33.50 & -- & -- \\
\modelicon{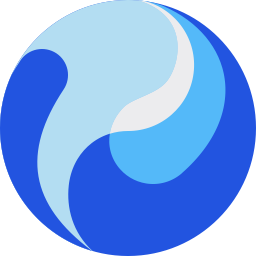} & Hunyuan-Turbos-Latest & 6.32 & 0.87 & 0.00 & 0.00 & 0.00 & 3.03 & 0.00 & 13.51 & 3.57 & 2.32 & 3,727.55 & 132.67 \\
\modelicon{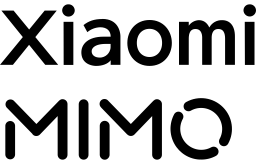} & Mimo-V2-Flash & 28.68 & 8.33 & 2.22 & 13.46 & 6.02 & 10.87 & 11.63 & 23.53 & 36.36 & 12.48 & 5,109.82 & 37.98 \\
\modelicon{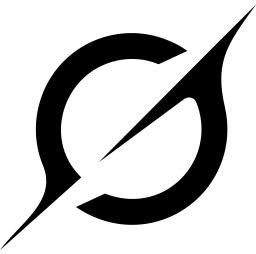} & Grok-4.1-Fast-Reasoning & 29.66 & 12.12 & 2.19 & 8.41 & 6.58 & 20.00 & 17.50 & 32.65 & 25.93 & 13.77 & 9,010.00 & 75.62 \\
\modelicon{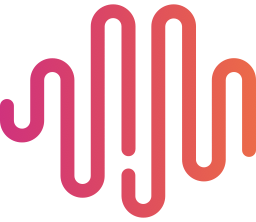} & MiniMax-M2.1 & 31.46 & 15.62 & 7.08 & 16.25 & 12.50 & 23.33 & 20.00 & 27.27 & 19.23 & 17.12 & 8,881.57 & 118.32 \\
\modelicon{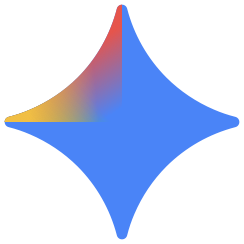} & Gemini-3-Flash & 32.76 & 16.89 & 4.10 & 14.95 & 10.60 & 17.95 & 18.18 & 30.61 & 41.38 & 17.62 & 6,563.28 & 50.56 \\
\modelicon{gemini} & Gemini-3-Pro-Preview & 61.98 & 20.83 & 1.71 & 26.74 & 19.11 & 13.64 & 28.57 & 52.00 & 55.56 & 27.52 & 5,815.14 & 80.80 \\
\modelicon{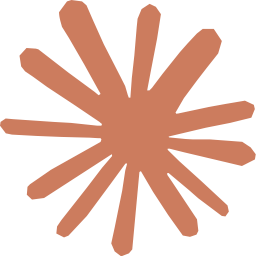} & Claude-Sonnet-4.5 & 68.22 & 14.86 & 1.79 & 16.13 & 22.30 & 29.27 & 23.81 & 47.73 & 44.83 & 26.36 & 8,586.84 & 91.43 \\
\modelicon{claude} & Claude-Opus-4.5 & 59.09 & 41.18 & 22.33 & 37.18 & 34.59 & 47.50 & 35.71 & 57.45 & 56.52 & 41.14 & 13,152.75 & 166.66 \\
\modelicon{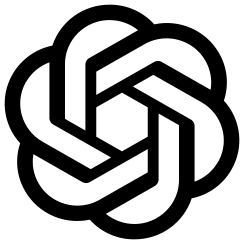} & GPT-5.1 & 74.71 & 21.37 & 3.49 & 24.14 & 18.10 & 33.33 & 45.83 & 57.78 & 64.71 & 32.00 & 11,256.15 & 154.09 \\
\midrule
\rowcolor{green!10}
\multicolumn{14}{c}{\textbf{Our Models}} \\
\anticon & Ling-3.0-Flash & 12.94 & 8.99 & 5.49 & 5.79 & 5.41 & 6.14 & 6.91 & 27.26 & 19.14 & 9.05 & -- & -- \\
\anticon & Ling-3.0-Flash + SFT & 37.41 & 22.45 & 22.44 & 24.17 & 16.04 & 31.58 & 25.53 & 50.00 & 53.12 & 26.85 & -- & -- \\
\anticon & Ling-3.0-Flash + RL (Ours) & 53.39 & 35.03 & 37.60 & 39.80 & 29.49 & 24.14 & 54.29 & 63.46 & 70.00 & \textbf{41.25} & -- & -- \\
\bottomrule
\end{tabular}%
}
\end{table*}

\paragraph{MiniAppBench.}
Table~\ref{tab:miniapp-main} shows substantial gains from training with the proposed environment. The base Ling-3.0-Flash model achieves an average pass rate of $9.05\%$, which increases to $26.85\%$ after SFT. Our RL framework further raises to \textbf{41.25\%}, improving upon SFT by 14.40 percentage points and the base model by $32.20$ percentage points. The resulting model surpasses Claude Opus 4.5 and GPT-5.1, achieving the highest average pass rate among the models in this comparison.

The improvement extends across all three difficulty levels. Relative to SFT, pass rates increase from $37.41\%$ to $53.39\%$ on easy tasks, from $22.45\%$ to $35.03\%$ on medium tasks, and from $22.44\%$ to $37.60\%$ on hard tasks. The model also improves in five of the six domains, reaching $39.80\%$ on Games, $29.49\%$ on Science, $54.29\%$ on Humanities, $63.46\%$ on Visualization, and $70.00\%$ on Lifestyle. Tools is the exception, decreasing from $31.58\%$ to $24.14\%$. These results demonstrate broad gains across application types and difficulty levels, while also revealing uneven benefits across domains.

\begin{table}[t]
\centering
\caption{\textbf{Results on ArtifactsBench.}
Reported leaderboard scores are shown alongside our SFT and RL models, ordered by AVG score. Highlighted rows denote our models.}
\label{tab:artifact-main}
\footnotesize
\setlength{\tabcolsep}{4pt}
\setlength{\abovecaptionskip}{3pt}
\setlength{\belowcaptionskip}{1pt}
\renewcommand{\arraystretch}{1.09}
\providecommand{\lblogo}[1]{\raisebox{-0.20em}{\includegraphics[height=1.1em,width=1.35em,keepaspectratio]{img/logo_official_png/#1.png}}}
\providecommand{\lbmodel}[2]{\lblogo{#1}\hspace{0.35em}#2}
\resizebox{\linewidth}{!}{%
\begin{tabular}{@{}clc@{}}
\toprule
\textbf{Rank} & \textbf{Model} & \textbf{AVG $\uparrow$} \\
\midrule
\rowcolor{green!15}
\textbf{1} & \textbf{\lbmodel{ant_ling}{Ling-3.0-Flash + RL (Ours)}} & \textbf{76.19} \\
2 & \lbmodel{openai}{GPT-5} & 72.55 \\
\rowcolor{green!8}
3 & \lbmodel{ant_ling}{Ling-3.0-Flash + SFT} & 71.71 \\
4 & \lbmodel{minimax}{MiniMax-M2} & 66.80 \\
5 & \lbmodel{claude}{Claude Opus 4.1} & 59.76 \\
6 & \lbmodel{ant_ling}{Ling-1T} & 59.31 \\
7 & \lbmodel{gemini}{Gemini 2.5 Pro} & 57.74 \\
8 & \lbmodel{claude}{Claude Sonnet 4} & 57.28 \\
9 & \lbmodel{openai}{GPT-OSS-120B} & 56.91 \\
10 & \lbmodel{qwen}{Qwen3-235B-A22B-Thinking} & 55.01 \\
11 & \lbmodel{openai}{o3-2025-04-16} & 54.04 \\
12 & \lbmodel{zhipu}{GLM-4.5} & 51.33 \\
\bottomrule
\end{tabular}%
}
\end{table}

\paragraph{ArtifactsBench.}
We next examine whether the gains extend to an external benchmark with different generation tasks and assessment criteria. As shown in Table~\ref{tab:artifact-main}, our RL model achieves a \textbf{state-of-the-art score of 76.19} on ArtifactsBench, ranking first in the overall leaderboard comparison. It improves upon the SFT initialization by $4.48$ points and surpasses GPT-5, the highest-scoring official leaderboard entry in this comparison, by $3.64$ points. These results demonstrate that the gains extend beyond MiniAppBench to another open-ended generation benchmark with a different evaluation protocol.

\subsection{Ablation Study}
\label{sec:ablation}

We examine the contributions of hierarchical rewards and diagnostic feedback through three RL configurations, all starting from the same SFT initialization. The \emph{binary rubric reward} uses standard GRPO with reward $1$ only when all functional rubrics are satisfied and $0$ otherwise. The \emph{hierarchical reward} retains standard GRPO but provides graded rewards based on format validity, source-code validity, runtime validity, and rubric-level fulfillment. The \emph{full framework} additionally uses diagnostic attributions to determine token-level optimization weights.

\begin{table}[t]
\centering
\caption{\textbf{Ablation and optimization analysis on MiniAppBench.}
All RL variants start from the same SFT initialization. Results are average pass rates over eight evaluation runs.}
\label{tab:ablation}
\footnotesize
\setlength{\tabcolsep}{3.5pt}
\setlength{\abovecaptionskip}{3pt}
\setlength{\belowcaptionskip}{1pt}
\renewcommand{\arraystretch}{1.13}
\begin{tabular}{@{}>{\raggedright\arraybackslash}p{0.36\linewidth}>{\raggedright\arraybackslash}p{0.41\linewidth}r@{}}
\toprule
\textbf{Variant} & \textbf{Feedback Used} & \textbf{Avg.} \\
\midrule
\rowcolor{black!4}
SFT & Supervised initialization & 26.85 \\
Binary reward & All-pass/all-fail rubric reward & 27.00 \\
Hierarchical reward & Graded validity and functionality rewards & 33.00 \\
\rowcolor{green!10}
Full framework & Hierarchical reward + diagnostic token weights & \textbf{41.25} \\
\midrule
\multicolumn{3}{@{}l}{\textit{Optimization analysis}} \\
Unconditional penalty & Hierarchical reward + unconditional error penalization & 37.00 \\
\bottomrule
\end{tabular}
\end{table}

\paragraph{Hierarchical rewards improve outcome discrimination.}
As shown in Table~\ref{tab:ablation}, the binary rubric reward achieves $27.00\%$, improving over SFT by only $0.15$ percentage points. Replacing it with the hierarchical reward raises the pass rate to $33.00\%$, a gain of $6.00$ percentage points. An all-or-nothing reward treats every application with an unmet requirement alike, obscuring differences in validity and task fulfillment. The hierarchical reward makes these differences available to group-relative optimization, providing useful learning signals before an application satisfies every requirement.

\paragraph{Diagnostic feedback strengthens policy optimization.}
Under the same hierarchical reward, incorporating diagnostic token weights further increases the pass rate from $33.00\%$ to $41.25\%$, a gain of $8.25$ percentage points. This comparison demonstrates the benefit of using the environment's findings to guide updates within each response, beyond using its overall assessments to compare responses. Together, the ablations support both roles of the proposed environment: distinguishing complex task outcomes and translating local diagnoses into effective optimization guidance.

\subsection{Analysis: Integrating Local Diagnoses with Group-Relative Learning}
\label{sec:diagnostic-analysis}

We further examine how diagnostic feedback should interact with the response-level advantage. A natural alternative is to penalize diagnosed error regions directly, regardless of the response's relative quality. We therefore compare the full framework with an \emph{unconditional error penalization} variant that uses the same hierarchical rewards and diagnostic localization but applies negative updates to diagnosed error regions in both positive and negative advantage responses.

As shown in Table~\ref{tab:ablation}, unconditional error penalization achieves $37.00\%$, falling below the full framework by $4.25$ percentage points. Local diagnoses and group-relative advantages convey different information: the former identify problematic regions, whereas the latter indicate whether a response outperforms its alternatives. A response can therefore contain diagnosed errors while still receiving a positive advantage. Unconditionally penalizing these regions may overrides the group-relative update direction.

Our framework preserves the advantage-determined update direction, emphasizing diagnosed regions in negative updates and unaffected regions in positive updates. Its gains over unconditional penalization support aligning local diagnostic feedback with group-relative learning.

\subsection{Transfer and Preservation of General Capabilities}
\label{sec:transfer}

We further examine whether training with the proposed environment preserves the model's broader capabilities while improving target-task performance. To examine this potential trade-off, we select benchmarks that probe complementary aspects of general capability: Aider-Polyglot for code editing across programming languages, Multi-IF for multi-turn multilingual instruction following, AA-LCR for long-context reasoning, AA-Omniscience for knowledge reliability, and AIME-26 for mathematical reasoning~\cite{aider_polyglot,multi_if,aa_lcr,aa_omniscience,aime2026}. Comparing the SFT and RL models on these tasks allows us to assess whether application-focused training preserves capabilities beyond the target domain and produces transferable benefits.

\begin{table}[t]
\centering
\caption{\textbf{Additional benchmark results.}
Scores follow their respective benchmark scales. The numerical change is computed as $\Delta=\mathrm{RL}-\mathrm{SFT}$.}
\label{tab:additional-benchmarks}
\footnotesize
\setlength{\tabcolsep}{4pt}
\setlength{\abovecaptionskip}{3pt}
\setlength{\belowcaptionskip}{1pt}
\renewcommand{\arraystretch}{1.10}
\begin{tabular}{@{}lrrr@{}}
\toprule
\textbf{Benchmark} & \textbf{SFT} & \textbf{RL} & \textbf{$\Delta$} \\
\midrule
\multicolumn{4}{@{}l}{\textit{Code and instruction following}} \\
Aider-Polyglot & 25.80 & 28.00 & +2.20 \\
Multi-IF & 74.30 & 75.95 & +1.65 \\
\addlinespace[1pt]
\multicolumn{4}{@{}l}{\textit{Reliability and reasoning}} \\
AA-LCR & 49.20 & 51.40 & +2.20 \\
AA-Omniscience & 18.67 & 20.50 & +1.83 \\
AIME-26 & 82.14 & 82.40 & +0.26 \\
% Verify whether this is the raw hallucination rate or a transformed score.
% Retain the supplied values, but do not interpret this change as a gain yet.
\bottomrule
\end{tabular}
\end{table}

Table~\ref{tab:additional-benchmarks} shows that performance on these benchmarks is maintained after RL, suggesting that the target-task gains do not come at the expense of the broader capabilities assessed here. Scores remain stable or improve modestly across code editing, instruction following, long-context reasoning, knowledge reliability, and mathematical reasoning, supporting capability preservation beyond the training domain.

Together with the ArtifactsBench results, these findings support a broader value of the proposed training environment: application-focused reinforcement learning can improve target-task performance while preserving, and in several cases strengthening, capabilities exercised in other settings. The gains are therefore consistent with learning transferable skills rather than merely adapting to a single benchmark.

\section{Conclusion}

We presented a reinforcement learning framework for open-ended tasks built around a training environment that integrates human domain knowledge with model reasoning. Rubrics anchor evaluation to task intent, hierarchical rewards distinguish multidimensional outcomes, and diagnostic attributions translate local findings into token-level optimization weights. Together, these mechanisms connect task specification, evaluation, and credit assignment within a unified training process.

Instantiated in interactive web application generation, the framework achieves strong results on MiniAppBench and ArtifactsBench. Ablations demonstrate the contributions of hierarchical rewards and diagnostic weighting, while further analysis highlights the importance of coordinating local diagnoses with group-relative learning. Additional evaluations support the preservation of broader capabilities alongside target-task gains. These findings suggest that training environments for open-ended tasks should provide both a basis for judging success and actionable guidance for policy optimization. Future work will extend this design to other open-ended domains and explore iterative refinement of evaluation criteria as policy capabilities evolve.

\appendix

\clearpage
\suppressfloats[t]
\section*{Appendix}

\lstset{
	basicstyle=\small\ttfamily,
	breaklines=true,
	columns=fullflexible,
	keepspaces=true,
	showstringspaces=false,
	frame=single
}

\section{Notation and Feedback Representation}
\label{app:formulation}

Let $x^{\mathrm{raw}}$ denote an original request and $x$ the expanded specification provided to the policy. The same specification is used to construct a rubric reference $\mathcal{R}(x)=\{r_1,\ldots,r_M\}$, which remains fixed across rollouts for that query. Conditioning on the shared generation instruction is implicit. In our web application instantiation, the generated response $y$ contains the HTML, CSS, and JavaScript implementing the requested application.

For each response, the environment returns a scalar reward and diagnostic findings,
\begin{equation}
	\mathcal{E}(x,y,\mathcal{R})
	=\bigl(R(y,\mathcal{R}),D(y)\bigr).
\end{equation}
The reward determines the response-level GRPO advantage, while the diagnostic findings identify unmet requirements and supporting implementation evidence. These findings are grounded in source spans and aligned with generated-token positions. The resulting weights modulate token contributions to the policy objective without replacing the response-level advantage.

This representation connects task specification, evaluation, and optimization through a shared reference. Different implementations can satisfy the same requirements, while deficiencies in individual responses can be associated with different implementation regions.

\section{Implementation of the Training Environment}
\label{app:environment}

\subsection{Query-Grounded Rubric Construction}
\label{app:rubric-construction}

Human domain knowledge enters the environment through shared rubric-construction principles, functional priorities, severity criteria, and reward rules. An LLM applies these specifications to individual queries through a five-stage pipeline comprising controlled query expansion, requirement extraction, functional module grouping, atomic decomposition, and final filtering. Table~\ref{tab:construction-stages} summarizes the pipeline, and Section~\ref{app:construction-prompts} provides the corresponding prompt templates.

\begin{table*}[t]
	\centering
	\small
	\caption{Stages of query-grounded rubric construction.}
	\label{tab:construction-stages}
	\begin{tabularx}{\textwidth}{clXX}
		\toprule
		Stage & Operation & Main input & Main output \\
		\midrule
		0 & Controlled expansion & Original query & Expanded query and expansion trace \\
		1 & Requirement extraction & Expanded query & Requirement map with identifiers, roles, and evidence \\
		2 & Module grouping & Query and requirement map & Initial-state and dynamic modules \\
		3 & Atomic decomposition & Query, requirements, and modules & Candidate items and decomposition decisions \\
		4 & Final filtering & Preceding construction outputs & Final reference, coverage report, and filtering actions \\
		\bottomrule
	\end{tabularx}
\end{table*}

Controlled expansion identifies the task's core objects, actions, states, outputs, and hard constraints, together with underspecified choices that could produce materially different task interpretations. High-risk choices with substantial reward impact are resolved using minimal, testable defaults. The expansion preserves the requested application and excludes unrelated features, external services, and implementation preferences. Its trace records alternative interpretations, frozen decisions, and causal relationships between actions, state changes, computations, and visible results.

The expanded query is exposed to the policy and serves as the sole task specification for subsequent rubric construction. A decision recorded only in the internal expansion trace cannot support a rubric requirement unless it is also expressed in the query. This shared specification reduces the risk that reward differences arise from evaluation criteria unavailable to the policy.

Requirement extraction produces an explicit map with stable identifiers, requirement types, functional roles, supporting query evidence, and a \texttt{must\_cover} indicator. Requirements are then grouped into user-observable modules, such as state systems, interaction workflows, validation categories, and output mechanisms. Module boundaries follow application behavior rather than DOM operations, variable names, or implementation steps. Covered requirement identifiers and query evidence connect each module to the task specification.

Atomic decomposition converts these modules into candidate items targeting independently failing states, rules, calculations, outputs, or interactions. Each candidate includes a \texttt{verification} description specifying the behavior or state to inspect. Final filtering merges duplicate or overlapping candidates, refines overly broad items, and removes unsupported requirements. Initial-state items concern page-load or no-action properties; dynamic items concern interaction-driven or runtime behavior. Ambiguities remaining in the expanded query are preserved rather than resolved by adding new requirements during decomposition.

The functional taxonomy distinguishes \texttt{core\_module}, \texttt{correct\_rule}, and \texttt{rendering\_effect}. These roles describe required capabilities and workflow surfaces, behavioral correctness conditions, and visible outputs or presentation effects, respectively. Mixed-role requirements and modules follow the precedence
\begin{multline}
	\text{\texttt{core\_module}}\succ
	\text{\texttt{correct\_rule}}\\
	\succ \text{\texttt{rendering\_effect}}.
\end{multline}
Mixed-role candidate items are split when possible; otherwise, the highest applicable role is retained. These are construction instructions applied by the model, rather than guarantees supplied by a deterministic classifier.

The enriched final reference contains seven fields per item: \texttt{id}, \texttt{requirement}, \texttt{verification}, \texttt{source\_module}, \texttt{covered\_requirement\_ids}, \texttt{query\_evidence}, and \texttt{rubric\_level}. A compact export retains the identifier, requirement, and level, with scope determined by membership in \texttt{initial\_state} or \texttt{dynamic}. The judge-facing payload makes scope explicit and exposes functional level as \texttt{level}. The final construction stage also requests coverage and faithfulness reports, uncovered requirements, unsupported items, and filtering decisions for inspection of the generated reference.

\subsection{Combined Code-Based Diagnosis}
\label{app:diagnosis}

We use \texttt{claude-opus-4.8} to produce three diagnostic layers in a single response: \texttt{static\_check}, \texttt{code\_simulation}, and \texttt{rubric\_alignment}. The judge assesses source validity, executability, and requirement satisfaction from the generated code. It does not execute browser interactions. Initial-state assessment examines initialization and rendering code, while dynamic assessment reasons about event handling, state transitions, computations, and their resulting outputs.

The static layer identifies HTML, CSS, and JavaScript syntax or lint issues. Critical findings indicate compile- or load-blocking syntax errors, major findings indicate non-blocking syntax or lint errors, and minor findings indicate low-impact suggestions. The simulation layer assesses executability independently of business correctness: critical findings indicate that the code effectively cannot run, major findings identify blocking points in an otherwise partially runnable implementation, and minor findings identify potential risks in runnable code. Critical or major findings fail the corresponding validity gate; minor findings alone do not.

The rubric layer compares the implementation with the query and reference. Critical findings denote entirely missing required features, major findings denote incorrect implementations that block the required interaction or make a feature unusable, and minor findings denote substantially implemented functionality with usability or presentation deficiencies. Minor findings pass the hard validity checks but retain small quality penalties in the continuous rubric reward.

Each diagnostic layer contains its name and three issue lists, with severity encoded by list membership. Each rubric finding identifies exactly one reference item and its scope. At most one finding is retained per identifier, using the highest applicable severity. A defect affecting several items produces separate item-level findings. The judge is instructed to assess every reference item but report only identified deficiencies, and to avoid penalizing requirements that are neither requested, implied, nor necessary to satisfy the query.

Findings can include exact code quotations, related symbols, implicated state variables, source-context hints, and an explanation of the evidence. Missing functionality is described as an absence; it does not justify inventing code quotations or treating an arbitrary insertion point as defective code. The diagnostic model returns structured findings, while numerical reward computation and source-coordinate resolution are performed programmatically. Section~\ref{app:diagnostic-prompt} specifies the consolidated diagnostic instruction and evidence schema.

\subsection{Source Localization and Contextual Expansion}
\label{app:localization}

Attribution combines model-generated evidence with programmatic source analysis. Issue normalization collects direct anchors from code quotations and error-subgraph fields, together with hints about relevant functions, state variables, event handlers, and related source regions. The implementation first attempts exact quotation matching, then whitespace-normalized matching, and finally matching smaller fragments separated by statement boundaries, braces, or line breaks. Matched anchors support local source windows extending approximately 1,200 characters on either side for contextual analysis.

A lightweight source analyzer extracts functions, arrow functions, and methods from script regions and records their character positions, approximate calls, state reads and writes, and event associations. Strings and comments are masked during analysis to reduce spurious matches. The resulting representation approximates selected code-property relations; it is not a complete program analysis or a full Joern implementation.

Expansion starts from the function enclosing a direct anchor. Implicated state variables connect the finding to regions defining, updating, or consuming that state. Caller and callee relations identify connected functions, while event bindings identify relevant interaction entry points. Rendering, scoring, and simulation regions are included when supported by the diagnostic evidence or these source relations. Duplicate regions are consolidated with their supporting reasons.

Direct quotations and expanded context remain distinct throughout attribution. A contextual window or enclosing function is not automatically treated as entirely defective. Direct token membership follows grounded direct spans, while contextual membership follows supported related regions. The trace preserves the source intervals, expansion relations, and HTML start offset required for token alignment. These relations indicate likely implementation responsibility; they do not establish unique causal responsibility through counterfactual execution.

\subsection{Token Alignment and Failure Handling}
\label{app:token-alignment}

Source attribution is mapped to generated-token positions using the tokenizer's character-offset mapping. Let $h_i$ denote the start of extracted HTML within response $y_i$. A source interval $[a,b)$ corresponds to $[h_i+a,h_i+b)$ in response coordinates. Token $t$, with nonempty character interval $[u_{i,t},v_{i,t})$, overlaps the source region when
\begin{equation}
	u_{i,t}<h_i+b
	\quad\text{and}\quad
	v_{i,t}>h_i+a.
	\label{eq:token-overlap}
\end{equation}
Alignment therefore depends on token positions, rather than token-ID values that may recur elsewhere in the response. Whitespace-normalized matches must be resolved back to the original source coordinates before this alignment.

The resulting direct and contextual token sets are denoted by $\mathcal{T}^{\mathrm{dir}}_i$ and $\mathcal{T}^{\mathrm{ctx}}_i$. Direct membership takes precedence where regions overlap. Their nominal markers are $3.0$ and $1.5$, respectively, with a baseline of $1.0$ elsewhere. The sign-dependent policy weights are defined in Section~\ref{app:optimization}.

The workflow prepends baseline weights for prompt positions and carries the concatenated array as per-token data. Weights are packed with the corresponding token sequences and aligned with prediction targets after the next-token shift. Prompt and padding positions are excluded from the policy objective, which is computed over generated response tokens.

The failure-handling protocol distinguishes evaluator failures from application failures. Malformed JSON, missing required fields, or invalid rubric identifiers trigger one corrective retry; persistent failures are handled as missing rewards. Repeated findings for one reference item are merged using the highest severity. When an anchor matches multiple locations, enclosing functions, related symbols, and nearby evidence support disambiguation. Unresolved anchors are excluded from direct attribution rather than assigned to every matching occurrence. Valid regions remain usable when other anchors fail. If findings are present but none can be localized, the scalar reward is retained and the response receives uniform token weights of $1.0$. A valid diagnosis with no findings instead yields empty attribution sets and follows the ordinary sign-dependent rule.

\section{Detailed Reward Computation}
\label{app:reward}
\label{app:reward-details}

The reward hierarchy distinguishes format validity, source validity, code-inferred executability, and rubric satisfaction. The format gate $F(y)$ checks whether the required HTML output can be extracted. For diagnostic layer $L$, let $\mathcal{I}^{L}_{\mathrm{crit}}$ and $\mathcal{I}^{L}_{\mathrm{maj}}$ denote its critical and major issue sets. The source and executability gates are
\begin{align}
	S(y)&=\mathbf{1}\!\left[
	|\mathcal{I}^{\mathrm{static}}_{\mathrm{crit}}|
	+|\mathcal{I}^{\mathrm{static}}_{\mathrm{maj}}|=0\right],\\
	E(y)&=\mathbf{1}\!\left[
	|\mathcal{I}^{\mathrm{simulation}}_{\mathrm{crit}}|
	+|\mathcal{I}^{\mathrm{simulation}}_{\mathrm{maj}}|=0\right].
\end{align}
Minor-only findings and empty issue lists pass these gates. The reward is
\begin{equation}
	R(y,\mathcal{R})=
	\begin{cases}
		0, & F(y)=0,\\
		0, & F(y)=1,\ S(y)=0,\\
		0.1, & F(y)=S(y)=1,\\
		& \text{and } E(y)=0,\\
		R_{\mathrm{rubric}}, & F(y)=S(y)=1,\\
		& \text{and } E(y)=1.
	\end{cases}
	\label{eq:app-gated-reward}
\end{equation}
The gates determine score precedence even though the diagnostic layers are returned together. A rubric-level business-logic error does not itself imply a source-validity or executability failure.

For responses passing all gates, let $\mathcal{V}(y,\mathcal{R})$ contain the distinct rubric items with critical, major, or minor findings. Each item has a reference-defined functional level $\ell_j$ and a judge-assigned severity $q_j$. Its penalty is $p_j=p_{\ell_j,q_j}$, using the nonnegative deduction magnitudes in Table~\ref{tab:reward-weights}. Each reference item contributes at most once after consolidation, with the highest applicable severity retained. Penalties are accumulated across initial-state and dynamic items without division by the total number of rubric items:
\begin{equation}
	P(y,\mathcal{R})=
	\min\left\{
	\sum_{r_j\in\mathcal{V}(y,\mathcal{R})}p_j,\,
	1-\epsilon
	\right\},
	\label{eq:app-rubric-penalty}
\end{equation}
where $\epsilon$ is a small positive constant satisfying $0<\epsilon<1$. Critical and major findings describe requirement failures, while minor findings contribute smaller deductions for quality deficiencies. The truncation ensures $1-P(y,\mathcal{R})>0$.

The rubric reward is obtained through the affine mapping
\begin{equation}
	R_{\mathrm{rubric}}=0.2+0.8\bigl(1-P(y,\mathcal{R})\bigr).
	\label{eq:app-rubric-reward}
\end{equation}
An application with no reported rubric deficiencies receives reward $1$. As penalties accumulate, the reward decreases until the penalty reaches $1-\epsilon$, yielding a minimum of $0.2+0.8\epsilon$. Thus, applications passing all validity gates receive rewards in $(0.2,1]$, above the rewards assigned to earlier-stage failures.

\begin{table}[t]
	\centering
	\small
	\caption{Per-item penalty magnitudes $p_{\ell,q}$, accumulated before the affine reward mapping.}
	\label{tab:reward-weights}
	\begin{tabular}{lccc}
		\toprule
		Functional level & Critical & Major & Minor \\
		\midrule
		\texttt{core\_module} & 0.35 & 0.25 & 0.08 \\
		\texttt{correct\_rule} & 0.25 & 0.12 & 0.04 \\
		\texttt{rendering\_effect} & 0.15 & 0.06 & 0.02 \\
		\bottomrule
	\end{tabular}
\end{table}

Rubric penalties determine response-level rewards; diagnostic token weights subsequently determine how the resulting advantages enter the policy objective.

\section{Data Construction and Training Details}
\label{app:training-details}

\subsection{Supervised Training Data Construction}
\label{app:sft-data-construction}

We construct 55,000 queries for supervised data preparation by combining functional requirements with diverse user contexts. Teacher models generate answers to these queries, providing demonstration targets for SFT. The resulting supervised checkpoint initializes reinforcement learning.

The capabilities targeted by construction are organized into 12 operational categories, summarized in Table~\ref{tab:capability-taxonomy}. These categories cover execution reliability, interaction, workflow completion, state management, domain logic, information processing, simulation, presentation, and requirement completeness. They provide a vocabulary for specifying task demands and characterizing implementation weaknesses. Categories are not mutually exclusive: a query can require several capabilities to operate together.

\begin{table*}[t]
	\centering
	\small
	\caption{Operational capability categories used in training-data construction.}
	\label{tab:capability-taxonomy}
	\begin{tabularx}{\textwidth}{lp{0.40\textwidth}X}
		\toprule
		ID & Capability category & Scope \\
		\midrule
		C01 & Runtime Safety and Reliable Rendering & Executability, initialization, and safe rendering \\
		C02 & Executable Controls and Accessible Interaction & Buttons, gestures, keyboard input, and accessible interaction \\
		C03 & End-to-End Workflow Completion & Complete workflows and visible outcomes \\
		C04 & State Transition and Temporal Coordination & Phases, turns, timers, and action ordering \\
		C05 & History, Persistence and Reversible Operations & Undo, redo, replay, and persistent state \\
		C06 & Domain Rules, Constraints and Boundaries & Rules, formulas, permissions, and edge cases \\
		C07 & Input Validation and Actionable Feedback & Validation, error handling, and user feedback \\
		C08 & Data Modeling, CRUD and Record Consistency & Structured records and data mutations \\
		C09 & Querying and Cross-View Synchronization & Search, filtering, and linked views \\
		C10 & Reactive Computation and Simulation & Parameter-driven computation and simulation \\
		C11 & Visual, Spatial and Animation Fidelity & Geometry, visualization, and animation \\
		C12 & Requirement and Deliverable Completeness & Module coverage and required deliverables \\
		\bottomrule
	\end{tabularx}
\end{table*}

Query construction separates functional content from its application context. We first identify the functional requirement space of the target task collection and represent complex tasks through fine-grained capability sets. These specify concrete interactions, state transitions, information operations, structural requirements, visualizations, output transformations, and domain mechanisms. We then sample personas with different occupations, expertise, motivations, and usage environments, conditioning query generation on both the capability set and the persona.

This composition allows the same capabilities to appear in different applications and narratives while preserving interaction complexity and task completeness. It increases diversity through changes in user needs and task composition rather than local paraphrasing alone. Each query remains traceable to its conditioning capabilities and persona. The operational taxonomy organizes the broad capability space, while individual requirement sets determine the behaviors requested by each query.

\subsection{Capability- and Reward-Guided Selection for RL}
\label{app:rl-query-selection}

After SFT, the target policy can generate basic frontend code but continues to struggle with complex logic, multi-step state changes, and user interactions. We select 3,626 queries from the 55,000-query collection based on these remaining weaknesses and the discriminative feedback obtained from sampled implementations. Capability relevance identifies what the policy still needs to learn, while reward discrimination indicates whether its current attempts provide useful comparisons for group-relative optimization.

The capability criterion emphasizes unreliable rules, incomplete workflows, inconsistent state updates, and missing connections between actions and visible results. The reward criterion examines whether sampled implementations attain different levels of validity or requirement satisfaction. Equal rewards produce zero centered advantages; difficulty alone therefore does not establish a query's immediate training value. Partially successful implementations can provide informative comparisons when their differences are reflected in the environment's feedback.

Selection is intended to concentrate RL on tasks that remain challenging while providing resolvable differences among current policy samples. Capability and context annotations additionally support consideration of task coverage, reducing the risk of concentrating learning on a narrow set of application patterns. These considerations describe the selection rationale without imposing a fixed category quota or a universal reward-variance threshold.

For RL, each selected query is paired with its expanded specification and fixed rubric reference. The policy receives the expanded query and generates fresh responses for evaluation. Teacher-generated SFT answers do not serve as RL rollouts. The final RL set contains 3,626 query--reference pairs.

\subsection{Policy Optimization}
\label{app:optimization}

For each query, we sample $G=8$ responses and compute group-relative advantages from their hierarchical rewards:
\begin{equation}
	A_i=\frac{R_i-\operatorname{mean}(R_1,\ldots,R_G)}
	{\operatorname{std}(R_1,\ldots,R_G)+\varepsilon_A}.
\end{equation}
No further advantage centering or standardization is applied. Direct and contextual attribution sets determine the nominal diagnostic weights
\begin{equation}
	\bar w_{i,t}=
	\begin{cases}
		3.0, & t\in\mathcal{T}^{\mathrm{dir}}_i,\\
		1.5, & t\in\mathcal{T}^{\mathrm{ctx}}_i,\\
		1.0, & \text{otherwise}.
	\end{cases}
\end{equation}

A response containing errors may still receive positive advantage relative to other group members. To avoid amplifying positive updates specifically at diagnosed regions, effective weights depend on the advantage sign:
\begin{equation}
	w_{i,t}=
	\begin{cases}
		\bar w_{i,t}, & A_i<0,\\
		1.0, & A_i\geq0,\ t\in\mathcal{T}^{\mathrm{dir}}_i\cup\mathcal{T}^{\mathrm{ctx}}_i,\\
		1.2, & A_i\geq0,\ t\notin\mathcal{T}^{\mathrm{dir}}_i\cup\mathcal{T}^{\mathrm{ctx}}_i.
	\end{cases}
	\label{eq:app-effective-weights}
\end{equation}
This rule also applies to valid diagnoses with empty finding sets. If findings exist but all localization attempts fail, the fallback in Section~\ref{app:token-alignment} instead sets every generated-token weight to one.

Let
\begin{equation}
	\rho_{i,t}=
	\frac{\pi_\theta(y_{i,t}\mid x,y_{i,<t})}
	{\pi_{\theta_{\mathrm{old}}}(y_{i,t}\mid x,y_{i,<t})}
\end{equation}
be the probability ratio between the current and sampling policies. With lower and upper clipping parameters $0.2$ and $0.28$, respectively, the policy loss is
\begin{align}
&\mathcal{L}_{\mathrm{policy}}
= -\frac{1}{G}\sum_{i=1}^{G}\frac{1}{T_i}\sum_{t=1}^{T_i}\nonumber\\
&\min\!\Bigl(\rho_{i,t}w_{i,t}A_i,\nonumber\\
&\operatorname{clip}(\rho_{i,t},0.8,1.28)\,w_{i,t}A_i
\Bigr).
\label{eq:app-policy-loss}
\end{align}
The objective averages over generated tokens within each response and then over responses. It is not normalized by the sum of diagnostic weights. Rewards, advantages, and attribution weights are fixed during each policy update. The value model is disabled, the KL coefficient is zero, and each sample is used once with one minibatch.

\subsection{Training Configuration and Evaluation}
\label{app:configuration}

Training uses 64 H200 GPUs across eight nodes. An AReaL-based stack uses SGLang for rollout generation and Megatron for policy optimization, with Ray execution and colocation enabled. Asynchronous training is disabled. The rollout allocation uses data parallelism 16 and tensor parallelism 4. The actor allocation uses data parallelism 8, tensor parallelism 2, and pipeline parallelism 4; expert parallelism is 8, expert tensor parallelism is 1, and context parallelism is 1.

The policy configuration has 42 layers, hidden size 2,560, 32 attention heads, and 512 routed experts with top-8 routing. Training uses BF16, distributed optimization, and activation recomputation. A batch contains 64 queries with eight responses per query, yielding a nominal 512 responses. Missing rewards are permitted, so the retained response count can be smaller. Table~\ref{tab:training-config} summarizes the training and generation settings.

\begin{table}[t]
	\centering
	\small
	\caption{Training and rollout configuration.}
	\label{tab:training-config}
	\begin{tabularx}{\linewidth}{Xl}
		\toprule
		Setting & Value \\
		\midrule
		SFT query pool & 55,000 \\
		Selected RL queries & 3,626 \\
		Hardware & 64 H200 GPUs \\
		Random seed & 42 \\
		Queries per batch & 64 \\
		Responses per query & 8 \\
		Nominal responses per batch & 512 \\
		Actor micro-batch size & 1 \\
		Maximum training updates & 2,000 \\
		Optimizer & Muon \\
		Learning rate & $4.5\times10^{-6}$ \\
		Learning-rate schedule & Constant after warmup \\
		Warmup iterations & 10 \\
		Weight decay & $0.01$ \\
		Gradient clipping threshold & $1.0$ \\
		Precision & BF16 \\
		Policy KL coefficient & $0$ \\
		Value model & Disabled \\
		Sample reuse & 1 \\
		Clipping interval & $[0.8,1.28]$ \\
		Sampling temperature & $1.0$ \\
		Top-$p$ & $1.0$ \\
		Top-$k$ & $1{,}000{,}000$ \\
		Maximum new tokens & 31,743 \\
		Total token limit & 32,767 \\
		Dataset \texttt{max\_length} & 2,048 \\
		Maximum concurrent rollouts & 256 \\
		Rollout timeout & 7,200 seconds \\
		Diagnostic model & \texttt{claude-opus-4.8} \\
		\bottomrule
	\end{tabularx}
\end{table}

Training queries are shuffled. The configured top-$k$ exceeds the vocabulary size of 157,184 and therefore introduces no additional truncation of the sampling distribution. Dataset and generation length limits are applied as separate constraints. The final checkpoint is selected by jointly inspecting the training reward and policy entropy trajectories.

MiniAppBench and ArtifactsBench evaluation follows the respective official benchmark procedures and scoring rules. Benchmark scores are obtained from their evaluation protocols, independently of the scalar reward used during training. The main experimental results, ablations, and additional evaluations are reported in the main text.

\section{Properties of the Reward and Weighted Objective}
\label{app:analyses}

The accumulated penalty is nonnegative and truncated at $1-\epsilon$. Consequently,
\begin{equation}
	\begin{aligned}
		0&\leq P(y,\mathcal{R})\leq1-\epsilon,\\
		0.2+0.8\epsilon&\leq R_{\mathrm{rubric}}\leq1.
	\end{aligned}
\end{equation}
For a fixed reference and unchanged existing findings, adding a finding cannot increase the reward. Before truncation, an additional penalty of $p_j$ reduces the final reward by $0.8p_j$. Once the accumulated penalty reaches the truncation threshold, further findings leave the scalar reward unchanged.

Each deduction depends on the item's functional level and severity rather than the total number of requirements. This preserves the same contribution for a given violation across references of different lengths, while allowing multiple deficiencies to accumulate. The construction-stage treatment of duplicate and overlapping items therefore matters: repeated reports for one identifier are consolidated, but semantically overlapping items with different identifiers could still yield separate deductions.

Diagnostic weighting scales the clipped policy term while preserving its clipping boundaries and advantage sign. For a fixed positive weight $w$,
\begin{align}
	&\min\bigl(\rho wA,\operatorname{clip}(\rho,0.8,1.28)wA\bigr)
	\nonumber\\
	&\quad=w\min\bigl(\rho A,\operatorname{clip}(\rho,0.8,1.28)A\bigr).
\end{align}
The sign-dependent rule increases the negative update emphasis on diagnosed regions in below-average responses. In above-average responses, diagnosed positions retain baseline weight while other positions receive weight $1.2$. Those other positions are not independently certified as correct; their larger coefficient expresses relative emphasis under the available diagnosis.

Because weights are not constrained to have unit mean, they affect both the distribution of emphasis within a response and its aggregate contribution to the objective. They cannot create a learning signal when all group rewards are identical: centered advantages are then zero, regardless of attribution. Likewise, multiplying a clipped term by a fixed positive weight does not restore a gradient where that term is locally flat. These properties explain why reward discrimination and diagnostic localization serve complementary roles.

\section{Prompt Templates}
\label{app:prompts}

\subsection{Rubric Construction}
\label{app:construction-prompts}

The publication templates below preserve the construction rules and output schemas while omitting redundant formatting instructions. Braced placeholders are populated with the query and preceding-stage outputs. The schemas use type placeholders such as \texttt{string}, \texttt{number}, and \texttt{boolean}; actual responses contain valid JSON values.

\begin{lstlisting}[caption={Stage 0: Controlled Query Expansion}]
	You are a structured MiniApp query canonicalization pipeline for GRPO training data.
	
	Goal:
	Rewrite the original query into a controlled-length canonical query. The rewrite may be longer than the original, but every added detail must reduce semantic variance inside a rollout group, so reward differences are more likely to reflect implementation quality rather than different task interpretations.
	
	Pipeline rules:
	1. Identify the MiniApp's core task, key objects, main user actions, required states, user-visible outputs, and hard constraints.
	2. Identify underspecified dimensions that could make rollouts implement different tasks, such as rule variants, input fields, update timing, output format, invalid actions, or completion conditions.
	3. Freeze only high-risk and high-reward-impact decisions. Use common, minimal, testable defaults.
	4. Model the causal chain from user action to state change, rule or calculation, and visible result.
	5. Remove or downgrade entropy-increasing wording such as optional, creative, beautiful, rich animation, dashboard, theme, or feel free unless explicitly required.
	6. Preserve the original app type, task, and hard constraints. Do not add unrelated product features, persistence, login, backend, external API, leaderboard, extra modes, AI opponents, export behavior, or technical-stack requirements.
	
	Return JSON only:
	{
		"expanded_query": string,
		"query_expansion_trace": {
			"core_task": {
				"primary_goal": string,
				"miniapp_type": string,
				"core_objects": [string],
				"main_actions": [string],
				"main_outputs": [string],
				"hard_constraints": [string]
			},
			"variance_sources": [
			{
				"dimension": string,
				"possible_branches": [string],
				"risk": "low" | "medium" | "high",
				"reward_impact": "low" | "medium" | "high",
				"why_it_matters": string
			}
			],
			"frozen_decisions": [
			{
				"decision": string,
				"value": string,
				"reason": string,
				"surface_in_query": boolean
			}
			],
			"causal_reward_anchors": [
			{
				"anchor": string,
				"action": string,
				"state_change": string,
				"rule_or_calculation": string,
				"visible_result": string,
				"surface_in_query": boolean
			}
			],
			"quality_audit": {
				"intent_preserved": boolean,
				"high_risk_variance_frozen": boolean,
				"unrequested_product_expansion": boolean,
				"rubric_or_reward_leakage": boolean,
				"group_consistency_risk": "low" | "medium" | "high",
				"over_expansion_risk": "low" | "medium" | "high"
			}
		}
	}
	
	Original query:
	{original_query}
\end{lstlisting}

\begin{lstlisting}[caption={Stage 1: Requirement Extraction}]
	Extract explicit MiniApp reward requirements from the expanded query only.
	
	Rules:
	- The expanded query is the only source of truth.
	- Do not add typical app behavior, rollout behavior, domain assumptions, persistence, export, AI behavior, validation, scoring, or implementation details unless explicitly stated.
	- Extract explicit UI groups, initial values or states, rules, calculations, flows, invalid actions, reset or end conditions, and required outputs.
	- Split independently checkable requirements while keeping natural UI groups together.
	- Each requirement must include non-empty query evidence from the expanded query.
	- Assign each requirement exactly one role:
	- core_module: essential task module, required input/control, data object, workflow surface, or page-load capability.
	- correct_rule: rule, formula, calculation, validation, state transition, interaction logic, or action-state-result behavior.
	- rendering_effect: visible output, visual feedback, chart, label, animation, highlighting, or presentation of computed/current state.
	- If a requirement fits multiple roles, choose the highest applicable role in this order:
	core_module > correct_rule > rendering_effect.
	
	Return JSON only:
	{
		"query_requirements": [
		{
			"requirement_id": "Q1",
			"section": "ui|initial_state|rules|flow|invalid_actions|end_condition|other",
			"type": "initial_ui|initial_value|core_interaction|state_transition|business_rule|calculation|validation|completion|reset|output|other",
			"text": "One explicit requirement.",
			"query_evidence": "The expanded-query phrase supporting this requirement.",
			"requirement_role": "core_module|correct_rule|rendering_effect",
			"must_cover": true
		}
		]
	}
	
	Expanded query:
	{expanded_query}
\end{lstlisting}

\begin{lstlisting}[caption={Stage 2: Functional Module Grouping}]
	Group the explicit requirement map into coarse functional modules.
	
	Rules:
	- Use only the expanded query and the explicit requirement map.
	- A module is a user-observable state system, rule system, validation category, interaction workflow, or output mechanism.
	- Do not create modules for DOM elements, CSS styles, variable names, event names, data structures, or implementation steps.
	- Every module must include covered requirement ids and query evidence.
	- Every module must have one module level: core_module, correct_rule, or rendering_effect.
	- Choose the module level from the covered requirements. If mixed, use:
	core_module > correct_rule > rendering_effect.
	- Put page-load or no-action modules in initial_state_modules.
	- Put runtime behavior, interactions, state updates, calculations, validation, and outputs in dynamic_modules.
	- Keep modules coarse; do not create final rubrics in this stage.
	
	Return JSON only:
	{
		"initial_state_modules": [
		{
			"module_id": "I-M1",
			"name": "short_snake_case_name",
			"requirement": "A concise coarse page-load requirement.",
			"covered_requirement_ids": ["Q1", "Q2"],
			"query_evidence": "The expanded-query phrase supporting this module.",
			"module_level": "core_module|correct_rule|rendering_effect"
		}
		],
		"dynamic_modules": [
		{
			"module_id": "D-M1",
			"name": "short_snake_case_name",
			"requirement": "A concise coarse dynamic requirement.",
			"covered_requirement_ids": ["Q3", "Q4"],
			"query_evidence": "The expanded-query phrase supporting this module.",
			"module_level": "core_module|correct_rule|rendering_effect"
		}
		]
	}
	
	Expanded query:
	{expanded_query}
	
	Explicit requirement map:
	{requirement_map}
\end{lstlisting}

\begin{lstlisting}[caption={Stage 3: Atomic Rubric Decomposition}]
	Convert functional modules into atomic candidate reward items.
	
	Rules:
	- Use only the expanded query and the explicit requirement map.
	- Atomic means one main state, rule, validation, output, or interaction behavior; one evidence body; user-observable behavior; and one independently failing condition.
	- Split modules that combine independent calculations, state transitions, validations, outputs, or action-derived rules.
	- Do not split into DOM operations, CSS details, event names, variable names, data structures, or algorithmic implementation steps.
	- Every candidate must include covered requirement ids and query evidence.
	- Every candidate must include one rubric level: core_module, correct_rule, or rendering_effect.
	- If a candidate mixes levels, split it when possible. If it cannot be split, choose:
	core_module > correct_rule > rendering_effect.
	- Preserve ambiguity when the query is ambiguous; do not choose unrequested implementation strategies.
	- Prefer high-signal reward items over exhaustive acceptance-test scripts.
	
	Return JSON only:
	{
		"decisions": [
		{
			"module_id": "D-M1",
			"atomic": false,
			"reason": "Why this module is or is not atomic.",
			"children": [
			{
				"requirement": "One candidate reward item.",
				"verification": "What code behavior or user-facing state should be checked.",
				"can_fail_independently": true,
				"covered_requirement_ids": ["Q3"],
				"query_evidence": "The expanded-query phrase supporting this candidate.",
				"rubric_level": "core_module|correct_rule|rendering_effect"
			}
			]
		}
		],
		"candidate_rubrics": {
			"initial_state": [
			{
				"id": "I-C1",
				"requirement": "...",
				"verification": "...",
				"source_module": "I-M1",
				"covered_requirement_ids": ["Q1"],
				"query_evidence": "...",
				"rubric_level": "core_module|correct_rule|rendering_effect"
			}
			],
			"dynamic": [
			{
				"id": "D-C1",
				"requirement": "...",
				"verification": "...",
				"source_module": "D-M1",
				"covered_requirement_ids": ["Q3"],
				"query_evidence": "...",
				"rubric_level": "core_module|correct_rule|rendering_effect"
			}
			]
		}
	}
	
	Expanded query:
	{expanded_query}
	
	Explicit requirement map:
	{requirement_map}
	
	Functional modules:
	{modules}
\end{lstlisting}

\begin{lstlisting}[caption={Stage 4: Final Rubric Filtering}]
	Filter candidate reward items into the final rubric used by the LLM reward judge.
	
	Rules:
	- Use only the expanded query, explicit requirement map, functional modules, and candidate rubrics.
	- Drop or rewrite anything without query evidence and valid covered requirement ids.
	- Keep user-observable code behavior.
	- Avoid implementation details, exact text, styling, DOM operations, variable names, libraries, or algorithms unless explicitly requested.
	- Merge duplicates and overlapping items.
	- Keep items that can pass or fail independently.
	- initial_state must contain only page-load or no-action requirements.
	- dynamic must contain only interaction or runtime behavior.
	- Every final item must include exactly:
	id, requirement, verification, source_module, covered_requirement_ids, query_evidence, rubric_level.
	- rubric_level must be one of:
	core_module, correct_rule, rendering_effect.
	- Compute coverage recall over must-cover requirements and faithfulness precision over final rubrics.
	
	Return JSON only:
	{
		"reference_pipeline": {
			"initial_state": [
			{
				"id": "I1",
				"requirement": "...",
				"verification": "...",
				"source_module": "I-M1",
				"covered_requirement_ids": ["Q1"],
				"query_evidence": "...",
				"rubric_level": "core_module|correct_rule|rendering_effect"
			}
			],
			"dynamic": [
			{
				"id": "D1",
				"requirement": "...",
				"verification": "...",
				"source_module": "D-M1",
				"covered_requirement_ids": ["Q3"],
				"query_evidence": "...",
				"rubric_level": "core_module|correct_rule|rendering_effect"
			}
			]
		},
		"coverage": {
			"coverage_recall": number,
			"faithfulness_precision": number,
			"covered_requirement_ids": ["Q1", "Q2"],
			"uncovered_requirements": [],
			"unsupported_rubrics": []
		},
		"filter_actions": [
		{
			"rubric_ids": ["D-C1", "D-C2"],
			"relation": "duplicate|overlap|dependent|independent",
			"action": "merge|keep|rewrite|drop",
			"reason": "..."
		}
		]
	}
	
	Expanded query:
	{expanded_query}
	
	Explicit requirement map:
	{requirement_map}
	
	Functional modules:
	{modules}
	
	Candidate rubrics:
	{decomposition}
\end{lstlisting}

\subsection{Combined Diagnosis and Evidence}
\label{app:diagnostic-prompt}

The consolidated instruction below specifies the diagnostic layers and evidence interface. It separates model-generated findings from programmatic reward computation and source localization. The input consists of the expanded query, the compact rubric payload, and the generated application source.

\begin{lstlisting}[caption={Combined diagnostic instruction and evidence schema}]
	You are a combined diagnostic agent for MiniApp generation.
	
	Perform exactly three diagnostic layers:
	1. static_check: HTML/CSS/JS syntax and lint quality.
	2. code_simulation: runtime executability independent of business correctness.
	3. rubric_alignment: implementation alignment with the query and reference.
	
	Read the supplied source code. Do not claim to have executed the application, operated a browser, or observed screenshots.
	
	Return one valid JSON object with exactly these top-level keys:
	static_check, code_simulation, rubric_alignment.
	
	Each layer must contain exactly:
	{
		"layer": "...",
		"critical_issues": [],
		"major_issues": [],
		"minor_issues": []
	}
	
	Do not output markdown or explanations outside JSON.
	Do not output weighted_score, final_reward, score, coverage, warnings, or additional top-level fields.
	Severity is determined by issue-list membership; do not add a severity field to issues.
	
	Issue objects must include type and message.
	For rubric_alignment, each issue must also include exactly one rubric_id copied from the reference and a scope of initial_state or dynamic.
	An evidence object may be included as specified below.
	
	Layer definitions:
	- static_check:
	critical: compile/load-blocking syntax errors.
	major: non-blocking syntax or lint errors.
	minor: low-impact style or syntax suggestions.
	Do not report business logic errors in this layer.
	
	- code_simulation:
	critical: the code effectively cannot run.
	major: the code is partially runnable but blocked at key points.
	minor: the code is runnable with potential risk points.
	Assess executability separately from task correctness.
	
	- rubric_alignment:
	critical: a required feature is completely missing.
	major: a required feature is implemented incorrectly so that the interaction cannot progress or the feature is unusable.
	minor: the feature is substantially implemented but has usability, polish, or experience deficiencies.
	
	Rubric rules:
	- The payload contains initial_state and dynamic items with id, scope, level, and requirement.
	- Assess all reference items and report only identified deficiencies.
	- If one implementation defect affects several reference items, return one separate issue per affected rubric_id.
	- Do not combine multiple rubric identifiers or unrelated errors in one issue.
	- Return at most one issue per rubric_id, using the highest applicable severity.
	- If an item is neither explicitly requested, implied by the query, nor necessary to satisfy the query, do not penalize its absence.
	- Return [] for every empty issue list.
	
	Evidence schema:
	{
		"code_quotes": [],
		"related_symbols": [],
		"state_variables": [],
		"related_regions": [],
		"location_hint": "",
		"rationale": ""
	}
	
	Evidence rules:
	1. code_quotes must be exact, contiguous quotations copied from the source. Prefer short quotations identifying the implementation associated with the issue.
	2. related_symbols must name functions, handlers, or identifiers present in the source.
	3. state_variables must identify state whose initialization, update, or use is relevant to the issue.
	4. related_regions may contain exact quotations showing relevant callers, callees, event bindings, state consumers, or rendering updates.
	5. location_hint may describe an enclosing function, event handler, or nearby context. rationale must explain how the evidence supports the issue.
	6. Do not invent code, identifiers, execution traces, or browser observations.
	7. For entirely missing functionality, explain the absence in the message. Leave code_quotes empty unless existing code directly supports the diagnosis.
	8. A possible insertion point is not evidence that existing code is defective.
	9. Use empty arrays or an empty string for unsupported evidence fields.
	10. Do not generate character offsets or token positions; these are resolved programmatically.
	11. Evidence for a rubric finding must support its single rubric_id.
	12. If an explicit diagnostic truncation marker is present, do not treat the artificial cut-off itself as a syntax error.
	
	Expanded query:
	{expanded_query}
	
	Rubric reference:
	{rubric_payload}
	
	Application source:
	{html_source}
\end{lstlisting}

\subsection{Policy Generation}
\label{app:policy-prompt}

The shared generation instruction accompanies the expanded query. Role delimiters are omitted below; the serialized assistant continuation begins with an empty \texttt{<think></think>} block.

\begin{lstlisting}[caption={Shared policy generation instruction}]
	You are an excellent web application designer and developer engineer, helping users complete web application development.
	
	The applications you develop aim to provide users with immersive experiences and help them acquire information and knowledge more efficiently through **interactive** web applications.
	**Important: Interactive means supporting users to actively change variables with corresponding result changes, not just simple content folding (like navigation bars) or content pagination.**
	
	Your current task is: Generate a runnable pure HTML + JavaScript page, output in plain text format.
	
	Output format requirements:
	Output complete HTML page code:
	- Use "#### `index.html`" marker
	- Followed by complete HTML code block (wrapped in ```html)
	- HTML code must include complete <!DOCTYPE html>, <html>, <head>, <body> tags
	- All CSS styles can be inline in <style> tag
	- All JavaScript code can be inline in <script> tag
	- Ensure the page is self-contained and can be opened directly in a browser to run
	
	Important notes:
	- **Interactivity requirements**:
	- Application must support users actively changing variables with corresponding result changes
	- Ensure the application provides true interactive experience
	
	- **Technical constraints**:
	- **Absolutely prohibit using fonts.googleapis.com, not accessible in some networks**
	- Ensure responsive design, adapt to different screen sizes
	
	- **Functionality implementation constraints**:
	- **Pure frontend implementation**, no backend API or external service calls
	- **Must provide mock data option**, ensure the application can run without user data
	- **AI functionality**: Do not use any AI API or LLM calls
	
	Please generate an HTML page according to the above constraints:
	detailed thinking off
\end{lstlisting}

\section{Worked Example of Rubric Evaluation and Attribution}
\label{app:case-study}

\subsection{A Pipeline-Generated Reference}
\label{app:q5}

We use the poker hand evaluator task, q5, to illustrate the connection between task requirements, diagnostic findings, and optimization weights. Its compact reference was extracted after the rubric construction pipeline. The task specification is reproduced below.

\begin{quote}
	\small
	Build a Poker Hand Evaluator miniapp. Display five random playing cards. The player must select the correct poker hand rank from the ten standard ranks (Royal Flush, Straight Flush, Four of a Kind, Full House, Flush, Straight, Three of a Kind, Two Pair, One Pair, High Card). Provide immediate visual feedback indicating correct or incorrect, revealing the correct rank if wrong. Include three difficulty modes: Easy (10 seconds per round, common hands), Medium (5 seconds per round, mixed hands), Hard (3 seconds per round, rare or ambiguous hands). Use a per-round countdown timer that auto-fails the round if it expires. Track cumulative score and session accuracy percentage. Display a static hand probability reference chart showing odds for each rank.
\end{quote}

Table~\ref{tab:q5-reference} reproduces the compact reference's identifiers, requirements, and stored functional levels. The reference contains five initial-state and twelve dynamic items. The displayed levels preserve the automatic assignments in this example rather than retrospectively relabeling them.

\begin{table*}[t]
	\centering
	\small
	\caption{Compact q5 reference. C, R, and E denote \texttt{core\_module}, \texttt{correct\_rule}, and \texttt{rendering\_effect}, respectively.}
	\label{tab:q5-reference}
	\begin{tabularx}{\textwidth}{lXc}
		\toprule
		ID & Requirement & Level \\
		\midrule
		\multicolumn{3}{l}{\textit{Initial-state requirements}} \\
		I1 & The initial UI must display five random playing cards. & R \\
		I2 & The initial UI must provide ten selectable poker hand rank options: Royal Flush, Straight Flush, Four of a Kind, Full House, Flush, Straight, Three of a Kind, Two Pair, One Pair, High Card. & R \\
		I3 & The initial UI must display a per-round countdown timer. & R \\
		I4 & The initial UI must provide three selectable difficulty modes: Easy, Medium, Hard. & C \\
		I5 & The initial UI must display a static hand probability reference chart showing odds for each rank. & R \\
		\midrule
		\multicolumn{3}{l}{\textit{Dynamic requirements}} \\
		D1 & Easy mode must use a 10-second countdown timer per round. & R \\
		D2 & Easy mode must generate common poker hands. & R \\
		D3 & Medium mode must use a 5-second countdown timer per round. & R \\
		D4 & Medium mode must generate mixed poker hands. & R \\
		D5 & Hard mode must use a 3-second countdown timer per round. & R \\
		D6 & Hard mode must generate rare or ambiguous poker hands. & R \\
		D7 & The player must be able to select a poker hand rank from the ten standard options. & R \\
		D8 & Provide immediate visual feedback indicating whether the selected rank is correct or incorrect. & E \\
		D9 & If the player selects incorrectly, reveal the correct poker hand rank. & E \\
		D10 & Auto-fail the round if the countdown timer expires. & R \\
		D11 & Track cumulative score across rounds. & R \\
		D12 & Track session accuracy percentage. & R \\
		\bottomrule
	\end{tabularx}
\end{table*}

The decomposition separates related but independently failing behaviors. Difficulty-dependent timing and hand generation are checked separately, as are a countdown display and its expiration behavior. Correctness feedback and revealing the correct answer also receive separate items. Qualitative descriptions such as ``common'' and ``rare or ambiguous'' remain as specified in the query; the reference does not introduce numerical sampling distributions for them.

\subsection{Diagnosis and Source Attribution}
\label{app:localization-example}

The following constructed implementation illustrates a failure of D12. It is a worked example of the attribution mechanism, not an observed training rollout. The required DOM elements and hand-selection functions are assumed to exist elsewhere, and the remaining application behavior is assumed to satisfy the reference.

\begin{lstlisting}[caption={Constructed accuracy-update defect}]
	<script>
	const state = { correct: 0, answered: 0 };
	
	function renderAccuracy() {
		const accuracy = state.answered === 0
		? 0 : 100 * state.correct / state.answered;
		document.getElementById("accuracy").textContent =
		accuracy.toFixed(1) + "%";
	}
	
	function submitAnswer(selectedRank, correctRank) {
		if (selectedRank === correctRank) {
			state.correct += 1;
			state.answered += 1;
		}
		renderAccuracy();
	}
	
	document.getElementById("submit").addEventListener("click", () => {
		submitAnswer(getSelectedRank(), getCorrectRank());
	});
	</script>
\end{lstlisting}

The denominator is updated only for correct submissions. Consequently, the accuracy display becomes 100\% after the first correct answer regardless of preceding incorrect answers. This is a functional correctness defect rather than a syntax or executability failure. An illustrative diagnostic response is:

\begin{lstlisting}[caption={Illustrative diagnostic output for D12}]
	{
		"static_check": {
			"layer": "static_check",
			"critical_issues": [],
			"major_issues": [],
			"minor_issues": []
		},
		"code_simulation": {
			"layer": "code_simulation",
			"critical_issues": [],
			"major_issues": [],
			"minor_issues": []
		},
		"rubric_alignment": {
			"layer": "rubric_alignment",
			"critical_issues": [],
			"major_issues": [
			{
				"type": "incorrect_accuracy_update",
				"rubric_id": "D12",
				"scope": "dynamic",
				"message": "The answered count increases only for correct answers, excluding incorrect submissions from session accuracy.",
				"evidence": {
					"code_quotes": [
					"if (selectedRank === correctRank) {\n    state.correct += 1;\n    state.answered += 1;\n  }"
					],
					"related_symbols": [
					"submitAnswer",
					"renderAccuracy"
					],
					"state_variables": [
					"state.correct",
					"state.answered"
					],
					"related_regions": [
					"const accuracy = state.answered === 0\n    ? 0 : 100 * state.correct / state.answered;"
					],
					"location_hint": "The correctness branch inside submitAnswer.",
					"rationale": "Incorrect submissions do not update the denominator consumed by renderAccuracy."
				}
			}
			],
			"minor_issues": []
		}
	}
\end{lstlisting}

The quoted branch grounds the direct anchor within \texttt{submitAnswer}. State-use relations connect this update to \texttt{renderAccuracy}, and the click listener identifies the interaction entry point. Table~\ref{tab:example-regions} summarizes these relations. Multiple relations supporting one region are consolidated without duplicating its token contribution.

\begin{table}[t]
	\centering
	\small
	\caption{Source relations supporting attribution in the worked example.}
	\label{tab:example-regions}
	\begin{tabularx}{\linewidth}{lX}
		\toprule
		Region & Supporting relation \\
		\midrule
		Quoted conditional & Direct implementation evidence for D12 \\
		\texttt{submitAnswer} & Function enclosing the direct anchor \\
		\texttt{renderAccuracy} & Consumer of implicated state and callee \\
		Submit click listener & Event entry invoking \texttt{submitAnswer} \\
		\bottomrule
	\end{tabularx}
\end{table}

\subsection{Reward and Token-Level Mapping}
\label{app:example-mapping}

Let the direct quotation occupy source interval $[a,b)$ and let $h$ be the HTML start offset within the generated response. Tokens overlapping $[h+a,h+b)$ receive direct membership; tokens overlapping only supported expanded regions receive contextual membership. An enclosing function is contextual evidence outside the directly quoted interval unless another direct anchor identifies additional positions.

The example passes the format, static, and executability gates under the stated assumptions. D12 has level \texttt{correct\_rule} and major severity, giving a single-item penalty of $0.12$. For the small $\epsilon$ regime where $1-\epsilon>0.12$, truncation is inactive and
\begin{equation}
	P=0.12.
\end{equation}
The corresponding rubric reward is
\begin{equation}
	R_{\mathrm{rubric}}
	=0.2+0.8(1-0.12)
	=0.904.
\end{equation}
The reference contains 17 items, but its size does not rescale this deduction. This is a worked calculation for the constructed implementation example.

The response advantage depends on the other responses in its rollout group. If $A_i<0$, direct, contextual, and remaining positions receive weights $3.0$, $1.5$, and $1.0$, respectively. If $A_i>0$, attributed positions retain weight $1.0$, while other positions receive weight $1.2$. Thus, the source attribution identifies the relevant implementation regions, and the group-relative advantage determines how their weights affect policy optimization.
\end{document}